%% file: frontierlab.tex
\documentclass[11pt, a4paper]{article}

\usepackage[utf8]{inputenc}
\usepackage[T1]{fontenc}
\usepackage{lmodern}
\usepackage{microtype}
\usepackage{graphicx}
\usepackage[section]{placeins}  
\usepackage{booktabs}
\usepackage{amsmath, amssymb, amsthm}
\usepackage{bm}
\usepackage{xspace}    
\usepackage{listings} 

\usepackage[a4paper, top=1.0in, bottom=1.1in, left=0.95in, right=0.95in, headheight=15pt]{geometry}
\usepackage{setspace}
\usepackage[skip=1em, indent=0pt]{parskip} 

\usepackage{xcolor}
\definecolor{seedblue}{RGB}{20, 56, 145}      
\definecolor{charcoal}{RGB}{30, 30, 32}       
\DeclareUnicodeCharacter{0300}{\`{}}
\definecolor{mutedgray}{RGB}{110, 110, 114}   
\definecolor{softrule}{RGB}{210, 210, 214}    

\usepackage[
    colorlinks=true,
    urlcolor=seedblue,
    linkcolor=seedblue,
    citecolor=seedblue,
    pdfborder={0 0 0}
]{hyperref}

\newcommand{\modelid}[1]{%
    \begingroup
    \Urlmuskip=0mu plus 1mu\relax
    \nolinkurl{#1}%
    \endgroup
}

\usepackage{titlesec}
\titleformat{\section}
    {\normalfont\large\bfseries\color{seedblue}}
    {\thesection}{0.8em}{}
\titlespacing*{\section}{0pt}{3.2ex plus 1ex minus .2ex}{1.4ex plus .2ex}

\titleformat{\subsection}
    {\normalfont\normalsize\bfseries\color{seedblue}}
    {\thesubsection}{0.7em}{}
\titlespacing*{\subsection}{0pt}{2.6ex plus 1ex minus .2ex}{1.0ex plus .2ex}

\titleformat{\subsubsection}
    {\normalfont\normalsize\itshape\color{seedblue}}
    {\thesubsubsection}{0.7em}{}

\titleformat{\paragraph}[runin]
    {\normalfont\normalsize\bfseries\color{seedblue}}
    {}{0pt}{}[.]
\titlespacing*{\paragraph}{0pt}{1.2ex plus .4ex minus .1ex}{0.7em}

\usepackage{fancyhdr}
\usepackage{lastpage}

\fancypagestyle{titlestyle}{%
    \fancyhf{}

    \fancyfoot[C]{\footnotesize\color{mutedgray}\thepage}
}

\newcommand{\see}{{\bfseries\color{charcoal}SEE}\xspace}

\usepackage[framemethod=TikZ]{mdframed}
\usepackage{thmtools}

\declaretheoremstyle[
    spaceabove=8pt,
    spacebelow=8pt,
    headfont=\normalfont\bfseries\color{charcoal},
    notefont=\normalfont\color{charcoal},
    notebraces={(}{)},
    bodyfont=\normalfont\itshape\color{charcoal},
    postheadspace=0.7em,
]{elegantstyle}

\usepackage[linesnumbered,ruled,vlined]{algorithm2e}
\SetAlFnt{\small}
\SetAlCapFnt{\normalfont\bfseries\color{charcoal}}
\SetAlCapNameFnt{\normalfont\bfseries\color{charcoal}}

\lstdefinestyle{promptstyle}{%
    basicstyle=\ttfamily\footnotesize\color{charcoal},
    backgroundcolor=\color{softrule!30},
    frame=single,
    rulecolor=\color{seedblue!60},
    framesep=6pt,
    xleftmargin=8pt,
    xrightmargin=8pt,
    breaklines=true,
    breakatwhitespace=true,
    columns=fullflexible,
    keepspaces=true,
    showstringspaces=false,
    aboveskip=8pt,
    belowskip=8pt
}

\usepackage{array}
\usepackage{tabularx}
\usepackage{longtable}
\usepackage{multirow}
\usepackage{makecell}
\usepackage{subcaption}
\usepackage{tikz}
\usetikzlibrary{positioning, arrows.meta}

\usepackage[
    labelfont={small,bf,color=charcoal},
    textfont={small,color=charcoal},
    labelsep=period,
    justification=justified,
    singlelinecheck=false
]{caption}

\usepackage{enumitem}
\setlist[itemize]{leftmargin=1.4em, itemsep=2pt, topsep=4pt, label={\small\textbullet}}
\setlist[enumerate]{leftmargin=1.6em, itemsep=2pt, topsep=4pt, label=\arabic*.}

\usepackage[
    style=numeric,
    sorting=none,
    maxnames=3,
    maxbibnames=99,
    backend=biber
]{biblatex}
\begin{document}

\thispagestyle{titlestyle}

\vspace*{-2cm}
{\setlength{\tabcolsep}{0pt}
\noindent
\raisebox{-0.5\height}{\includegraphics[height=0.92cm]{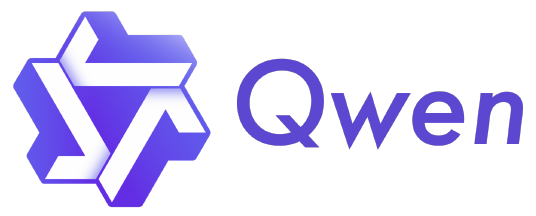}}%
\hspace{0.45cm}%
\raisebox{-0.5\height}{\includegraphics[height=0.80cm]{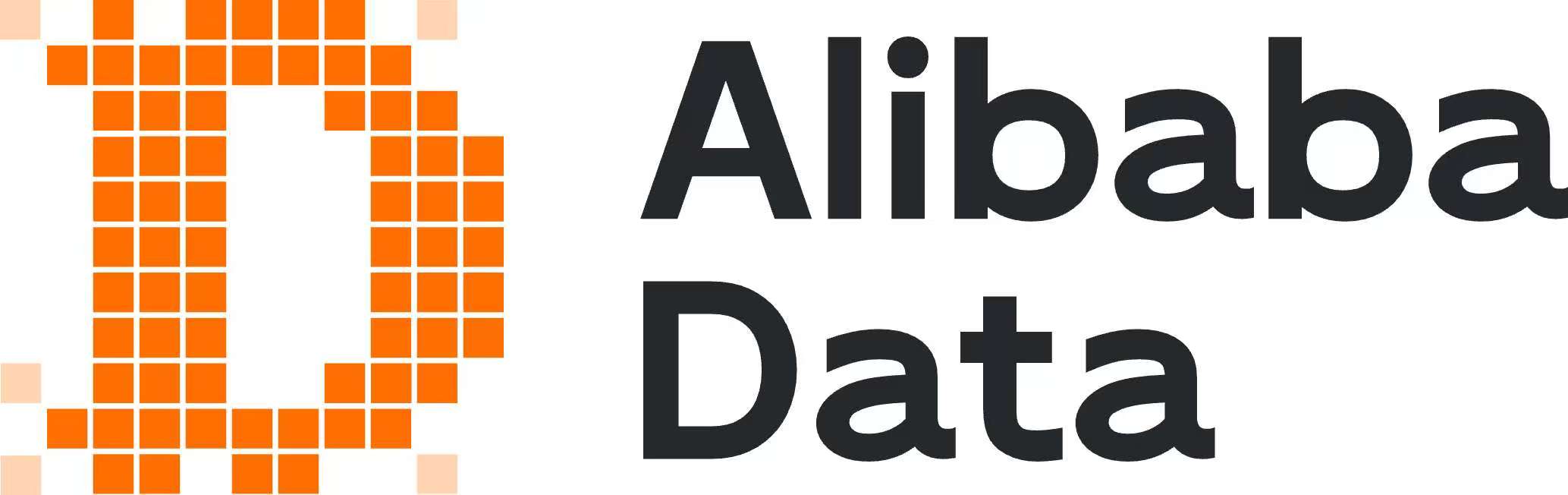}}%
\hfill
\raisebox{-0.5\height}{\includegraphics[height=0.7cm]{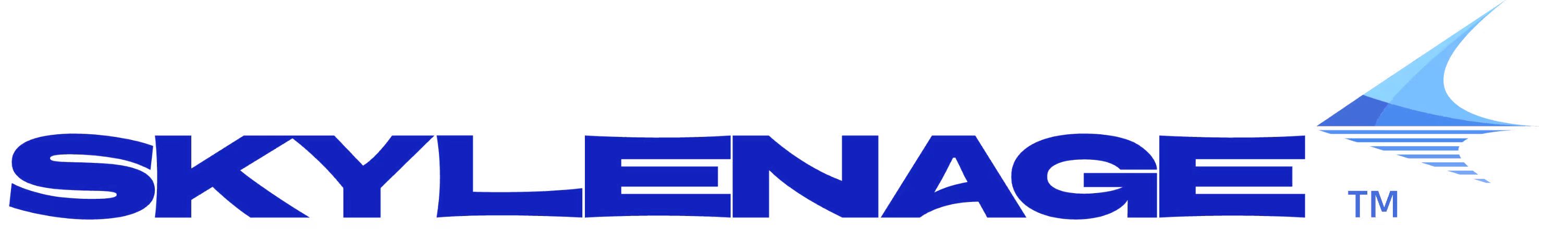}}%
\par}

\vspace{0.05cm}

\noindent{\color{seedblue}\rule{\textwidth}{0.6pt}}

\vspace{0cm}

\begin{center}

    {\fontsize{18}{22}\selectfont\bfseries\color{charcoal}%
    Science Edge Evaluation: SEE the Missing Step Toward Real Scientific Discovery\par}
\end{center}

\vspace{0cm}

\noindent{\color{seedblue}\rule{\textwidth}{0.6pt}}

\vspace{0.0cm}

\begin{center}
    {\normalsize\bfseries\color{charcoal}
    Taolin~Han\textsuperscript{1,3*},
    Yuchen~Zhang\textsuperscript{1,4*},
    Jinghang~Wang\textsuperscript{1*},
    Yun~Wu\textsuperscript{1},
    Wai~Yuet~Chiu\textsuperscript{1},
    Zhaohai~Li\textsuperscript{2},
    Yifei~Zhang\textsuperscript{1,5},
    Jinxin~Wang\textsuperscript{1,4},
    Yuhao~Zhou\textsuperscript{1,6},
    Chen~Zhao\textsuperscript{1,4},
    Jiajia~Li\textsuperscript{1},
    Jiaxin~Li\textsuperscript{1},
    Qile~Jin\textsuperscript{1},
    Kewei~Sun\textsuperscript{1},
    Shuang~Wu\textsuperscript{1},
    Weiqi~Zhai\textsuperscript{1},
    Renquan~Lv\textsuperscript{1,6},
    Junchao~Li\textsuperscript{1},
    Ruodan~Chen\textsuperscript{1},
    Qingteng~Chen\textsuperscript{1},
    Zhibo~Yang\textsuperscript{2},
    Hu~Wei\textsuperscript{1},
    Lin~Qu\textsuperscript{1},
    Shuai~Bai\textsuperscript{2,\dag},
    Bing~Zhao\textsuperscript{1,\dag}\par}

    \vspace{0.2em}

    {\small\color{charcoal}%
    \textsuperscript{1}Alibaba Group,\quad
    \textsuperscript{2}Qwen Team, Alibaba Group,\quad
    \textsuperscript{3}University of Chinese Academy of Sciences,\quad
    \textsuperscript{4}Tsinghua University,\quad
    \textsuperscript{5}University of Alberta,\quad
    \textsuperscript{6}Zhejiang University\par}

    \vspace{0.2em}

    {\footnotesize\color{mutedgray}
    \textsuperscript{*}Equal Contribution,
    \textsuperscript{\dag}Correspondence\par
     \href{mailto:wangjinghang.wjh@alibaba-inc.com}{wangjinghang.wjh@alibaba-inc.com}
    \par}

    \vspace{0.5em}
    {\small\color{charcoal}
    \href{https://github.com/SKYLENAGE-AI/science-edge-evaluation}{\raisebox{-0.2em}{\includegraphics[height=1em]{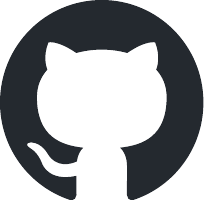}}~Code}\par}
\end{center}

\vspace{0.0cm}

\begin{mdframed}[
    linewidth=0.7pt,                
    linecolor=seedblue,             
    backgroundcolor=white,          
    roundcorner=6pt,                
    innertopmargin=22pt,            
    innerbottommargin=20pt,         
    innerleftmargin=28pt,           
    innerrightmargin=28pt,          
    skipabove=10pt,                 
    skipbelow=16pt                  
]
\begin{center}
    {\fontsize{18}{22}\selectfont\bfseries\color{seedblue} Abstract}
\end{center}
\vspace{0.3em}

\input{sections/00_abstract}
\end{mdframed}

\vspace{0.3cm}


\input{sections/01_introduction}
\input{sections/02_related_work}
\input{sections/03_method}
\input{sections/04_experiments}
\input{sections/05_discussion}

\section*{Data Availability}
To facilitate the benchmarking and reproducibility of our work, the accompanying code and public datasets are available on \href{https://github.com/SKYLENAGE-AI/science-edge-evaluation}{GitHub}. The main benchmark contains 1{,}116 questions, of which 1{,}049 are publicly released. To support reproducibility, we also report headline results on the public subset in Supplementary Tables~\ref{tab:si-public-subset} and~\ref{tab:si-public-subset-tool}.

\newpage
\section*{References}
\addcontentsline{toc}{section}{References}
\printbibliography[heading=none]


\newpage
\appendix
\setcounter{figure}{0}
\setcounter{table}{0}
\renewcommand{\thefigure}{S\arabic{figure}}
\renewcommand{\thetable}{S\arabic{table}}
\begin{center}
    {\fontsize{16}{20}\selectfont\bfseries\color{seedblue} Supplementary Information}
\end{center}
\vspace{0.5em}
\input{sections/07_supplementary}

\end{document}

%% file: sections/00_abstract.tex
{\setstretch{1.15}
\noindent
Large language models (LLMs) are increasingly involved in scientific discovery, yet it remains unclear whether they can support complex real laboratory science. Here we introduce {\bfseries\color{charcoal}Science Edge Evaluation} (\see), a multimodal benchmark of expert-curated questions grounded in peer-reviewed literature and experimental practice in chemistry, biology, and materials science. Evaluation of 28 multimodal large language models (MLLMs) shows that even the best-performing model reaches only 54.9\% accuracy. Moreover, general-purpose models outperform science-specialized models on average. In the visual-agent evaluation, the use of tools increases the best accuracy to 56.1\%. Tool use can expand the information available to models, but more information does not necessarily lead to reliable scientific reasoning. The key challenge is whether models can manage tool-derived information within the boundaries of the original experimental evidence. Together, these findings reveal that current MLLMs still cannot reliably make justified and evidence-bounded inferences from experimental results, which is an essential capability in real scientific discovery. Bridging this gap requires MLLMs to transition from explaining established scientific concepts to deriving novel and evidence-based insights from experimental data.
\par}

%% file: sections/01_introduction.tex
\section{Introduction}
\label{sec:intro}

Large language models (LLMs) and their agentic extensions are increasingly used as assistants for scientific discovery~\cite{ref:ai4science, ref:biochemllm_survey, ref:biomni}. Current evaluations frame science capability along either disciplinary knowledge or procedural research skill. The former uses text-only suites to score knowledge breadth over curated closed-ended items~\cite{ref:mmlu, ref:mmlupro, ref:cmmlu, ref:scibench}. Expert-level and domain-specific tests raise the difficulty of those items with frontier academic examinations and problems drawn from literature~\cite{ref:gpqa, ref:hle, ref:chembench, ref:macbench}, while Olympiad and exam-grade collections push the difficulty even further~\cite{ref:olympiadbench, ref:mmscibench, ref:mmesci, ref:usncov}. The latter requires agentic systems to assist or execute research workflows, from tool-augmented chemical experimentation and autonomous laboratories to end-to-end paper production~\cite{ref:chemcrow, ref:coscientist, ref:sdlab, ref:co_scientist, ref:aiscientist, ref:agent_laboratory}. Their evaluations reward procedures, including whether a plan was executed, a tool call was successful, or a workflow ended with an answer~\cite{ref:agentbench}. Even benchmarks that claimed to be built closer to laboratory practice mainly ask whether a domain-specific answer is correct~\cite{ref:labbench, ref:chemvts, ref:matbench, ref:polyreal}. High scores in these protocols certify knowledge coverage and procedural fluency~\cite{ref:contamination}, but do not certify the reasoning process that research depends on~\cite{ref:processreward}.

The distinction at issue is not one of difficulty but of what the task demands. In the laboratory, conclusions are assembled from imperfect observations, including spectra, gel images, or diffraction patterns, together with tabular measurements and textual context. Interpreting such evidence correctly is an expert skill~\cite{ref:figureintegrity}. A critical judgment, moreover, lies in determining which conclusions the observations can support and where the limits of that support lie. Existing protocols bypass this judgment. When an item is scored only against a single verified answer, a correct and evidence-bounded conclusion is indistinguishable from one that is correct despite overreaching reasoning. LLMs assert unsupported content fluently with poor calibration~\cite{ref:hallucination, ref:truthfulqa, ref:calibration}. What remains unmeasured, in both static and agentic settings, is the evidentiary boundary, i.e. whether a model reasons within the limits of the data it was given.

To close this gap, we introduce Science Edge Evaluation (\see), a multimodal benchmark constructed from authentic scientific tasks in chemistry, biology, and materials science. Its questions are adapted from peer-reviewed literature and experimental practice such that every answer must be licensed by the observations provided; \see thus scores not only whether a model reaches the correct conclusion, but also whether that conclusion remains within the bounds of the available evidence. Our evaluation of 28 MLLMs reveals that current models remain unreliable in realistic experimental settings and that their failures cannot be attributed to limited knowledge access alone. Rather, \see exposes a fundamental deficiency in evidence-based scientific reasoning. A complementary analysis of tool-augmented visual agents further shows that tool access expands evidence acquisition but does not obviate the need for reliable evidence management across the interaction trajectory. In short, current models struggle to ``SEE'' multimodal observations across disciplines while maintaining rigorous adherence to the evidence at hand. By aligning evaluation with laboratory practice, \see measures not only whether models possess scientific knowledge, but whether they can derive justified evidence-bounded conclusions from multimodal experimental data.

%% file: sections/02_related_work.tex
\section{Related Work}
\label{sec:related}

\paragraph{General Academic Benchmarks}
Academic benchmarks are essential for evaluating LLM and MLLM capabilities. General academic benchmarks such as MMLU~\cite{ref:mmlu}, MMLU-Pro~\cite{ref:mmlupro}, and GPQA~\cite{ref:gpqa} evaluate models across broad academic domains with different levels of difficulty. The tested capabilities include general academic reasoning, scientific question answering, mathematical reasoning, and code generation. Region-specific extensions such as CMMLU~\cite{ref:cmmlu} further broaden the evaluation surface. To challenge models at the frontier of expert-level reasoning, benchmarks such as HLE~\cite{ref:hle} have been specifically designed to incorporate closed-ended questions of exceptional difficulty.

\paragraph{Multimodal Benchmarks}
Multimodal benchmarks such as ScienceQA~\cite{ref:scienceqa} and MMMU~\cite{ref:mmmu} represent meaningful advances by extending evaluations beyond text-only paradigms through the incorporation of visual inputs covering scientific and academic subjects, with MMMU-Pro~\cite{ref:mmmupro} further removing text-solvable questions to enforce true multimodal reasoning. Complementary efforts including MM-Vet~\cite{ref:mmvet}, MathVista~\cite{ref:mathvista}, and SciFIBench~\cite{ref:scifibench} target integrated multimodal capabilities, mathematical reasoning in visual contexts, and scientific figure interpretation, respectively. Olympiad- and contest-grade benchmarks such as OlympiadBench~\cite{ref:olympiadbench}, the Chinese-oriented MMSciBench~\cite{ref:mmscibench}, the multilingual MME-SCI~\cite{ref:mmesci}, and USNCO~\cite{ref:usncov} built from chemistry-olympiad exams push difficulty further with bilingual or multilingual multimodal scientific problems. While current benchmarks are rooted in general academic and exam-style contexts, they fail to capture the experimental data and workflow-driven reasoning fundamental to authentic scientific inquiry.

\paragraph{Scientific Benchmarks}
Some recent benchmarks are designed to evaluate models in more specialized scientific settings, including chemistry, biology, and materials science. These efforts focus on domain-specific reasoning rather than broad academic knowledge, exemplified by ChemBench~\cite{ref:chembench}, SciBench~\cite{ref:scibench}, SciEval~\cite{ref:scieval}, and LAB-Bench~\cite{ref:labbench}. A few of them incorporate multimodal inputs such as figures, molecular structures, spectra, and tables, together with experimental results. For example, MaCBench~\cite{ref:macbench} is a multimodal benchmark for chemistry and materials science that is closer to real research scenarios, while Matbench~\cite{ref:matbench} targets materials property prediction. More recent efforts further probe modality-specific limitations, for example, ChemVTS-Bench~\cite{ref:chemvts} disentangles visual, textual, and symbolic chemical reasoning. Existing scientific benchmarks lack the disciplinary breadth, experimental realism, and diagnostic depth required to evaluate MLLMs against the complex, interdisciplinary challenges of authentic scientific research.

\paragraph{Scientific LLMs and Agents}
Scientific LLMs have been developed through continued pretraining, instruction tuning, and domain adaptation on scientific corpora. Representative examples include general scientific and biomedical models such as Galactica~\cite{ref:galactica}, BioGPT~\cite{ref:biogpt}, BioMedLM~\cite{ref:biomedlm}, GatorTronGPT~\cite{ref:gatortrongpt}, Med-PaLM~\cite{ref:medpalm}, and Meditron~\cite{ref:meditron}; chemistry-, molecule-, and materials-oriented models or instruction resources such as ChemLLM~\cite{ref:chemllm}, ChemDFM~\cite{ref:chemdfm}, Mol-Instructions~\cite{ref:molinstructions}, LlaSMol~\cite{ref:llasmol}, HoneyBee~\cite{ref:honeybee}, and MatterChat~\cite{ref:matterchat}; and multimodal biomedical or scientific models such as LLaVA-Med~\cite{ref:llava_med}, Med-PaLM M~\cite{ref:medpalm_m}, BioMedGPT~\cite{ref:biomedgpt}, Intern-S2~\cite{ref:model_interns2_official}, Intern-S2 Preview-35B~\cite{ref:model_interns2}, Intern-S2 Preview-397B~\cite{ref:model_interns2_397b}, and Intern-S1 Pro~\cite{ref:model_interns1}. Recent scientific agents further extend LLMs from static question answering to tool-augmented and workflow-level scientific assistance, including ChemCrow~\cite{ref:chemcrow}, Coscientist~\cite{ref:coscientist}, The AI Scientist~\cite{ref:aiscientist}, Agent Laboratory~\cite{ref:agent_laboratory}, and Co-Scientist~\cite{ref:co_scientist}. While these efforts demonstrate the growing role of LLMs in scientific workflows, existing evaluations often focus on knowledge recall, domain-specific task performance, tool-use success, or final-answer accuracy. In contrast, \see evaluates whether MLLMs can ground conclusions in multimodal experimental evidence, integrate concepts across disciplines, and reason within the boundaries of available evidence.

%% file: sections/03_method.tex
\section{Method}
\label{sec:method}

We organize \see through a staged construction and validation workflow, from real-world data sourcing and expert question design to AI-assisted quality control, expert review, and final benchmark acceptance (Figure~\ref{fig:see_overview}a).

\subsection{Data Collection}
\label{sec:method:collection}

\see is a collaborative effort. The questions are contributed by active frontline experts who hold master's or PhD degrees or are PhD candidates in chemistry, biology, materials science, or related interdisciplinary areas. The resulting benchmark spans three broad disciplines and 17 reported sub-fields, including both single-discipline and cross-disciplinary questions (Figure~\ref{fig:see_overview}b).

\paragraph{Sources}
The questions are drawn from real scientific research settings. Sources include figures and data from peer-reviewed papers, recent research literature, and first-hand experimental data contributed by experts. Visual evidence covers common experimental data structures in laboratory research, including bioactivity measurements, Cryo-EM structures, Western blot and gel electrophoresis images, spectra such as infrared spectroscopy, nuclear magnetic resonance (NMR) and mass spectrometry (MS), microscopy images such as SEM, TEM and AFM, X-ray diffraction patterns, and thermal analysis curves.

\paragraph{Style}
The questions in \see include choice-based, short-answer, numerical, measurement, and image-processing tasks. Some questions contain explicit option lists, while others do not use a separate option field and are evaluated through canonical short answers or numerical answers. The dataset also records task-type metadata, including reasoning, measurement, and image-processing categories. Each entry contains text and associated visual files. These associated images may include both visual evidence presented with the question and images used for expert verification. As a result, the reported image counts are not the exact number of figures displayed in the question itself. Each entry is reported with a standard answer to support automatic evaluation and expert verification.

\paragraph{Labeling}
Each question in \see is assigned discipline labels spanning biological sciences, chemistry, and materials science. The reported analysis uses a final taxonomy of 17 fine-grained labels covering all 1{,}116 questions. This labeling scheme captures cross-disciplinary questions clearly. Label distributions are reported in Supplementary Information~\ref{sec:si-eval-coverage}.

\subsection{Data Processing}
\label{sec:method:review}

A multi-stage review process is used to ensure data quality, difficulty, and reliability. After collection and label assignment, the questions that do not meet our criteria are filtered out.

\paragraph{Standardization}
The raw questions collected are standardized in terms of wording, terminology, answer structures, option numbering, numerical precision, tolerance records, and image organization. The images are normalized in format and resolution.

\paragraph{General Quality Control}
After standardization, questions are checked for adherence to the required format, clear and objective answers, sufficient visual and textual evidence, and correct labels. We remove semantic duplicates and questions that do not fulfill the requirements. Each question is then cross-checked by at least two experts in the relevant domain.

\paragraph{Expert Quality Control}
After AI-assisted checks, experts review the questions in terms of scientific accuracy. Questions with problems are returned to their contributors for revision. The revised questions then undergo the entire standardization and quality control process before inclusion.

\paragraph{Image-Ablation Checking}
Finally, questions with images are checked to verify whether images are indispensable for solving the task. Questions that can be answered from text alone are removed or revised. This verification is consistent with previous multimodal benchmark practices~\cite{ref:mmmu, ref:scifibench}.

\paragraph{Question Evaluation}
The evaluation of questions follows the standard scoring protocols established in large-scale evaluation frameworks~\cite{ref:helm}. For multiple choice questions, MLLM answers must \emph{exactly} match the ground-truth answers (including single- and multiple-response questions). Numerical fill-in-the-blank questions are evaluated with tolerance level provided by experts when applicable. Non-numerical fill-in-the-blank questions are evaluated against standardized canonical answers.

Representative examples of the questions collected are shown in Supplementary Figure~\ref{fig:data_composition}, illustrating within-discipline and cross-disciplinary multimodal scientific questions.

\subsection{Model Evaluation Environment}
\label{sec:method:evaluation_protocol}

We evaluate 28 representative MLLMs with \see, including 24 general-purpose models and four science-specialized models. The general-purpose models include Gemini 4 Argon~\cite{ref:model_gemini4}, GPT-6 Astra (Max)~\cite{ref:model_gpt6astra}, GPT-6.1 Sol (Max)~\cite{ref:model_gpt61}, GPT-5.6-Sol (Max)~\cite{ref:model_gpt56sol}, Gemini 3.1 Pro~\cite{ref:model_gemini3pro}, Claude Opus 5 (Max)~\cite{ref:model_claudeopus5}, Kimi K3.1 Pro~\cite{ref:model_kimi31}, Qwen3.8-Max~\cite{ref:model_qwen38max}, GPT-5.5 (xhigh)~\cite{ref:model_gpt55}, Kimi K3~\cite{ref:model_kimik3}, Gemini 3.5 Flash~\cite{ref:model_gemini3flash}, Grok 4.7~\cite{ref:model_grok47}, Qwen3.8 Flash~\cite{ref:model_qwen38flash}, Claude Opus 4.8 (Max)~\cite{ref:model_claudeopus48}, DeepSeek V4.1 Flash 1017~\cite{ref:model_deepseek41}, Seed2.1 Pro~\cite{ref:model_doubao21}, Qwen3.7-Plus~\cite{ref:model_qwen37}, Claude Opus 5.5 (Max)~\cite{ref:model_claude55}, Seed2.0 Pro~\cite{ref:model_doubao20}, MiniMax-M3~\cite{ref:model_minimaxm3}, Kimi K2.6~\cite{ref:model_kimik26}, MiMo V2.6 Pro~\cite{ref:model_mimo26}, GLM-5V-Turbo~\cite{ref:model_glm5v}, and MiMo-V2.5~\cite{ref:model_mimo}. The science-specialized models include Intern-S2~\cite{ref:model_interns2_official}, Intern-S2 Preview-35B~\cite{ref:model_interns2}, Intern-S2 Preview-397B~\cite{ref:model_interns2_397b}, and Intern-S1 Pro~\cite{ref:model_interns1}. Only four science-specialized models are tested because few models in this category possess the multimodal capabilities required for our evaluation.

Unless otherwise specified, models are evaluated with their recorded default inference configurations. Configurations labeled Max designate extended-thinking runs, and the GPT-5.5 row retains its recorded \texttt{xhigh} designation. For GPT-6 Astra, GPT-6.1 Sol, and Claude Opus 5.5, both default and Max configurations were evaluated and we report only the Max results. Detailed settings and the limits of recovered configuration records are provided in Supplementary Information~\ref{sec:si-eval-setup}. For image-containing questions, we apply the preprocessing required by each model interface, including format conversion, resizing, and input organization. We introduce no manual prompt correction, per-question parameter tuning, or result filtering. Claude models have refusal or safety-filtered no-response cases; under the strict binary scoring protocol, these instances are counted as incorrect.

Model outputs are evaluated with a strict binary LLM-as-a-judge pipeline~\cite{ref:llmjudge}. Each response is extracted and marked as correct or incorrect by comparing its final answer with the reference answer while allowing equivalent formats. Single- and multiple-choice answers must exactly match the reference option set, and numerical answers must match the reference value or fall within an expert-specified range or tolerance when one is provided. Refusals, non-answers, generation or judging failures, and responses from which no final answer can be identified are counted as incorrect, so all 1{,}116 questions remain in the denominator. Evaluation and judging prompts are detailed in Supplementary Information~\ref{sec:si-eval-prompt}, and judge sensitivity is assessed in Supplementary Information~\ref{sec:si-judge-sensitivity}.

\paragraph{Visual ablation}
For all 28 models, we compare image-conditioned and text-only accuracy on the same 1{,}116 question UUIDs while preserving the original textual queries. We additionally classify explicit acknowledgments that necessary visual evidence is missing. Accuracy and acknowledgment rates are interpreted separately because the generation prompt requests an answer rather than a calibrated abstention decision. Input consistency, paired results, and the acknowledgment criterion are detailed in Supplementary Information~\ref{sec:si-eval-textonly}.

\subsection{Tool-augmented visual-agent evaluation}
To evaluate whether tool access improves scientific reasoning beyond the standard multimodal setting, we further test eleven models in a tool-augmented visual-agent setting. Gemini 4 Argon supports web search; the other ten models support both web search and a code interpreter. Each model receives the same question text, options, and image inputs as in the standard evaluation, but may iteratively invoke its available tools before producing a final answer. Tool use is optional rather than forced, and no curated retrieval corpus, local database, or domain-specific scientific tool is provided. Final answers are scored with the same binary judging protocol as the standard evaluation. Accuracy comparisons use the main leaderboard no-tool baselines, although interfaces may differ between runs. For correctness-changing pairs, trajectory attribution uses the original paired controls, which may differ from the leaderboard runs. An attribution model first proposes mechanisms and supporting trace evidence; domain experts then review the original responses and tool outputs and revise the attributions as needed. Detailed configurations, attribution criteria, and tool-failure categories are provided in Supplementary Information~\ref{sec:si-tool-augmented-agent-eval}.

%% file: sections/04_experiments.tex
\section{Results and Discussion}
\label{sec:evaluation}

\subsection{Current MLLMs show limited reliability on SEE}
\label{subsec:quantitative}

\begin{figure}[!htbp]
    \centering
    \includegraphics[width=\textwidth]{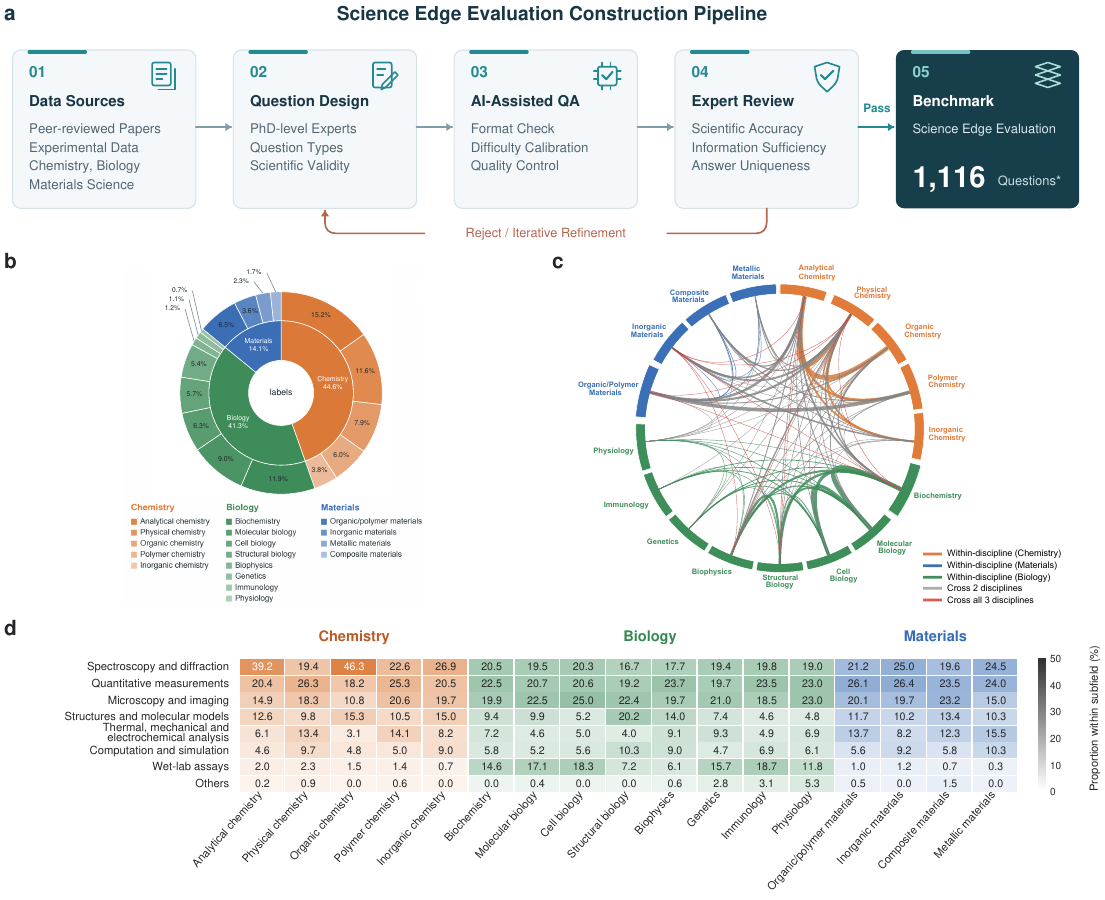}
    \caption{Overview of \see construction and composition.
    \textbf{a}, Construction pipeline for the 1{,}116-question benchmark. Questions were generated from literature or real experimental scenarios, standardized with AI-assisted checks, reviewed by domain experts and collected after quality-control criteria were met.
    \textbf{b}, Shares of the 2{,}895 sub-field memberships across three disciplines (inner ring) and 17 sub-fields (outer ring).
    \textbf{c}, Co-occurrence of sub-field labels within questions. Edge thickness is proportional to the number of shared questions. Within-discipline links use discipline colors; gray and red links mark questions spanning two and three disciplines, respectively.
    \textbf{d}, Evidence categories inferred from keywords in the question text and knowledge-point labels. Each question contributes equal fractions to its matched categories, or to Other when none match; each sub-field column sums to 100\% before rounding (Supplementary Information~\ref{sec:si-dataset}).}
    \label{fig:see_overview}
\end{figure}

\see contains 1{,}116 expert-curated questions derived from real scientific tasks in chemistry, biology and materials science, with 1{,}049 questions publicly released (Figure~\ref{fig:see_overview}a). The benchmark covers 17 reported sub-fields across three disciplines (Figure~\ref{fig:see_overview}b) and includes both single-discipline and cross-disciplinary tasks (Figure~\ref{fig:see_overview}c). Each question pairs textual context with multimodal experimental evidence, including figures, spectra, microscopy images, tables, measurement results and other laboratory data (Figure~\ref{fig:see_overview}d); representative examples are shown in Supplementary Figure~\ref{fig:data_composition}. This design allows us to evaluate whether MLLMs can reason from experimental evidence rather than rely only on scientific knowledge or text-solvable cues.

\begin{figure}[!htbp]
    \centering
    \includegraphics[width=\textwidth]{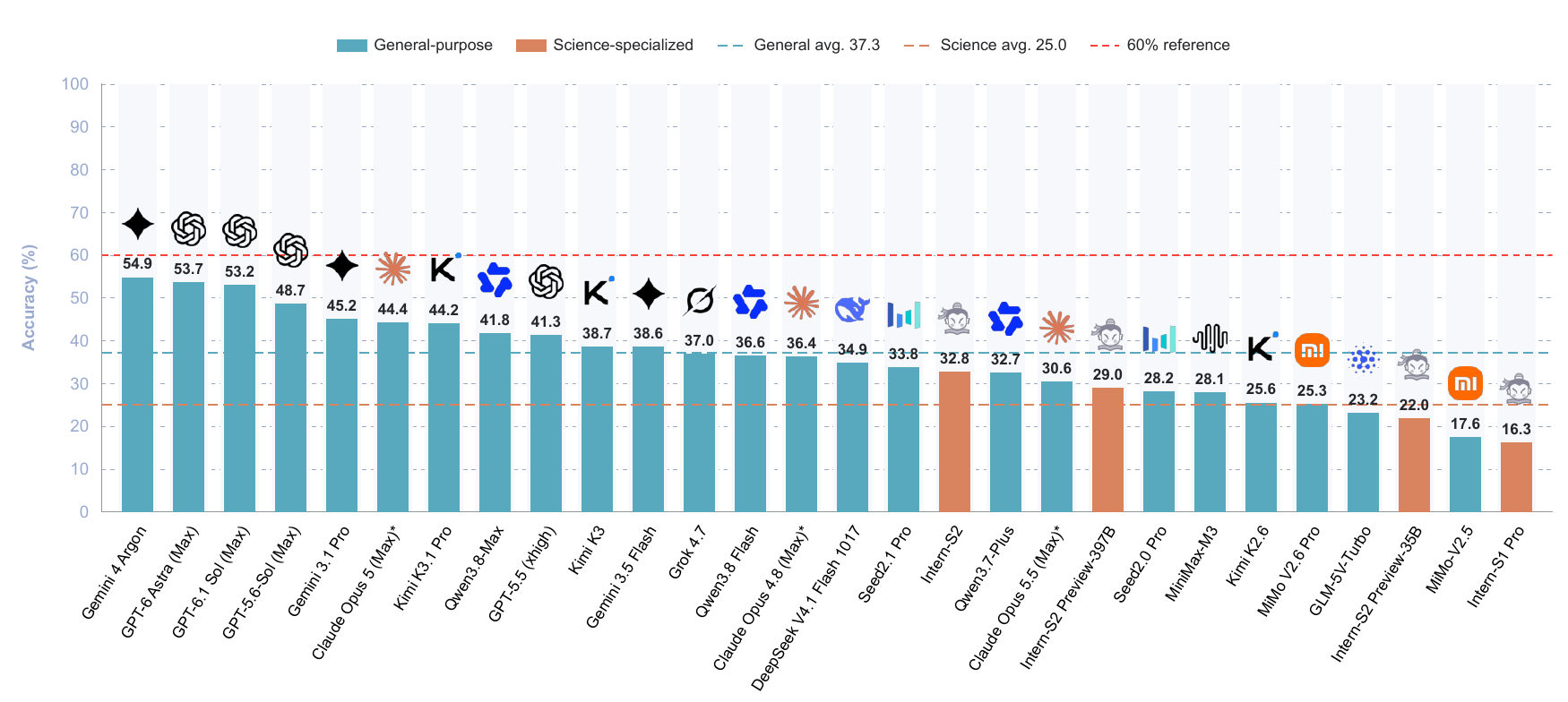}
    \caption{Overall accuracy of 28 MLLMs on \see. Teal and orange bars denote general-purpose and science-specialized models, respectively. Matching dashed horizontal lines indicate group means, and the red line marks a 60\% reference. Asterisks mark models with refusals or safety-filtered non-responses counted as incorrect.}
    \label{fig:model_performance}
\end{figure}

Following the \hyperref[sec:method:evaluation_protocol]{evaluation protocol described in the Methods}, we first report the overall accuracy of 28 representative MLLMs on \see. All evaluated models achieve low accuracy in the standard evaluation setting (Figure~\ref{fig:model_performance}). The best model, Gemini 4 Argon, reaches 54.9\% accuracy, followed by GPT-6 Astra (Max) at 53.7\% and GPT-6.1 Sol (Max) at 53.2\%. Across the 28 models, accuracy ranges from 16.3\% to 54.9\%, and no model exceeds 55\%.

The result at the discipline level shows similar aggregate precision in all three main disciplines (Tables~\ref{tab:accuracy_discipline} and~\ref{tab:accuracy_discipline_second} in Supplementary Information~\ref{sec:si-eval-finegrained}). Across all evaluated models, average accuracy is 35.6\% for chemistry, 34.7\% for biology, and 31.8\% for materials science.

Performance differences become clearer at the sub-discipline level (full per-model breakdown in Tables~\ref{tab:accuracy_discipline} and~\ref{tab:accuracy_discipline_second}). The lowest mean accuracies occur in organic and polymer materials (27.8\%), inorganic chemistry (30.5\%), polymer chemistry and physics (31.3\%), and composite materials (31.3\%). The highest reported sub-discipline accuracy is organic chemistry (41.6\%), followed by analytical chemistry (38.1\%), metallic materials (37.6\%), and genetics (37.0\%). Sub-disciplines with small sample sizes should be interpreted with care.

The per-model profiles show that no single model demonstrates universal dominance throughout the scientific spectrum, and proficiency remains highly domain-contingent. Claude models have refusal or safety-filtered no-response cases on a subset of biochemistry questions involving virus--host interactions, pathogen structural biology, and related experimental techniques. These cases are counted as incorrect under the strict evaluation protocol and marked with asterisks in the related figures and tables.

\subsection{General-purpose models outperform science-specialized models}
We next compare general-purpose and science-specialized MLLMs to examine whether domain adaptation improves performance on \see. Within the tested model set, the four science-specialized models remain below the general-purpose average on \see (Figure~\ref{fig:model_performance}), with an average accuracy of 25.0\% compared to 37.3\% for the 24 general-purpose models. However, Intern-S2 Preview-397B reaches 29.0\%, outperforming several mid-tier general-purpose models and narrowing the gap relative to earlier science-specialized models.

We observe that the phenomenon of leading general-purpose models outperforming domain-specific models is not unique to our setting, but also appears in other fields such as clinical medicine~\cite{ref:clinicallm}. This suggests that simply specializing a model through post-training or augmenting it with retrieval-augmented generation (RAG) does not necessarily lead to improved performance.

These results suggest that success on \see requires more than domain-specific factual familiarity. Science-specialized training may improve exposure to scientific terminology, concepts, and task formats, but \see requires models to combine such knowledge with visual experimental evidence and specific experimental context. The gap between science-specialized and leading general-purpose models therefore indicates that robust scientific reasoning in real experimental settings depends not only on domain adaptation, but also on broader multimodal understanding and flexible evidence-grounded reasoning. This observational comparison does not isolate specialization from differences in model scale, training, or inference configuration.

\subsection{Interdisciplinary complexity reduces model performance}
As noted above, greater exposure to a specific scientific domain does not necessarily translate into reliable reasoning across broader scientific contexts. In practice, scientific discovery often crosses disciplinary boundaries and requires integrating knowledge, assumptions, and observations from multiple fields. To reflect this reality, \see explicitly includes interdisciplinary questions spanning biology, chemistry, and materials science. This design is central to the benchmark, whose goal is not only to test whether models know isolated scientific facts, but also to evaluate whether they can integrate heterogeneous evidence across disciplinary contexts.

We first examine the performance gap between single-discipline and cross-discipline questions. Across all 28 models, the average accuracy is 38.6\% on the 602 questions within a single main discipline, but drops to 32.0\% on the 514 questions that span multiple main disciplines. This 6.6-percentage-point gap suggests that cross-disciplinary settings impose additional coordination demands beyond those captured by single-discipline evaluation.

We next examine cross-subdisciplinary complexity. Averaged across all models, performance remains similar for questions annotated with one or two subdisciplinary labels, with accuracies of 39.5\% and 39.3\%, respectively. However, the accuracy drops to 32.7\% for questions with three labels and further to 30.5\% for questions with four labels. The nine five-label questions yield 42.1\% accuracy, but their small sample provides limited support for generalization. This pattern suggests that performance often degrades as a task requires coordination across a larger number of scientific concepts, experimental techniques, or forms of evidence, although label count does not isolate task difficulty or establish a causal effect.

Together, these results show that interdisciplinary tasks provide a stringent stress test of model robustness. Such tasks require models to integrate scientific terminology, experimental assumptions, visual evidence, and multistep reasoning within a single response. The interdisciplinary design of \see therefore assesses whether current MLLMs can coordinate knowledge and evidence across the boundaries commonly encountered in real-world research. To be considered robust in a research setting, a model must maintain coherent reasoning even when a problem falls outside the distributions most frequently represented in training.

\subsection{Visual ablation exposes weak evidence-boundary awareness in MLLMs}
\begin{figure}[htbp]
    \centering
    \includegraphics[width=0.9\textwidth]{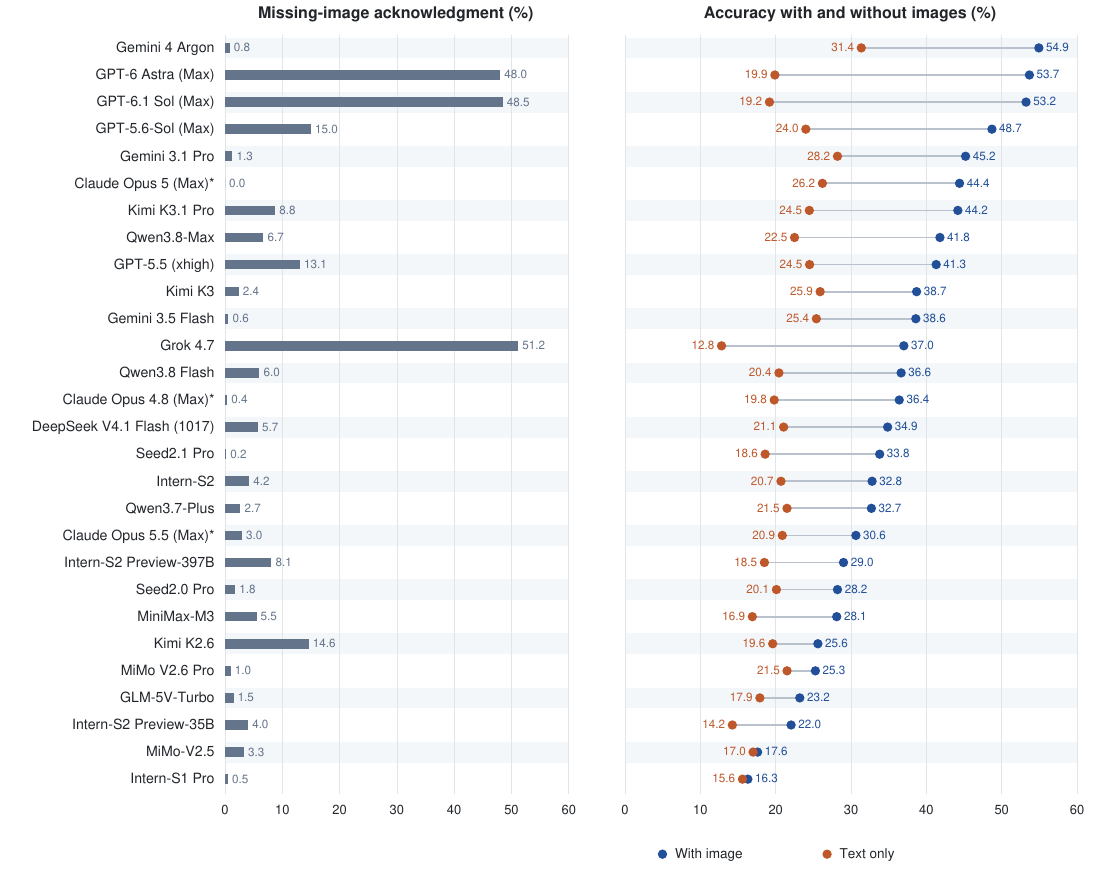}
    \caption{Image removal and missing-image acknowledgment across 28 models. Blue and orange dots show accuracy with and without images on the same 1{,}116 question UUIDs within each model. Gray bars show the percentage of text-only responses that explicitly cite missing visual evidence and withhold a concrete answer. Models are ordered by with-image accuracy. Empty outputs, generation errors, generic refusals, and other non-answer cases are not counted as acknowledgments. Asterisks mark safety-filtered non-answers scored as incorrect.}
    \label{fig:text_only_missing_image}
\end{figure}

To evaluate the level of hallucination of MLLMs by investigating their ability to recognize missing information, we performed a blind-modality diagnostic by removing image inputs while preserving the original textual queries across 28 models. This experiment serves as a critical test for grounding. A truly intelligent system should identify when a scientific conclusion is impossible without visual evidence. Across 31{,}248 text-only instances, models explicitly identified missing information in 2{,}888 cases (9.2\%, Figure~\ref{fig:text_only_missing_image}). This low rate suggests that many current MLLMs do not explicitly recognize missing empirical evidence, although acknowledgment rates vary widely and lack of acknowledgment alone does not establish hallucination because textual cues remain available and the generation prompt requests an answer. Removing the image lowers accuracy for every model, by 14.5 percentage points on average; further text-only diagnostics and input-version caveats are reported in Supplementary Information~\ref{sec:si-eval-textonly}.

This result reveals limited awareness of evidential boundaries in current MLLMs. When visual inputs are removed, models often continue to answer based on textual cues, prior knowledge, or common scientific patterns, rather than recognizing that the available evidence is insufficient. This behavior is especially concerning in real-world scientific settings, where decisions are often made under incomplete or ambiguous information and where unsupported but confident answers may pose greater risks than appropriate abstention. Thus, this ablation analysis provides a diagnostic signal that current MLLMs do not yet consistently reason within the limits of experimental evidence.

\input{sections/05a_error_analysis}

\subsection{Tool-augmented visual agents shift the bottleneck to evidence management}
\begin{figure}[!htbp]
    \centering
    \includegraphics[width=\textwidth]{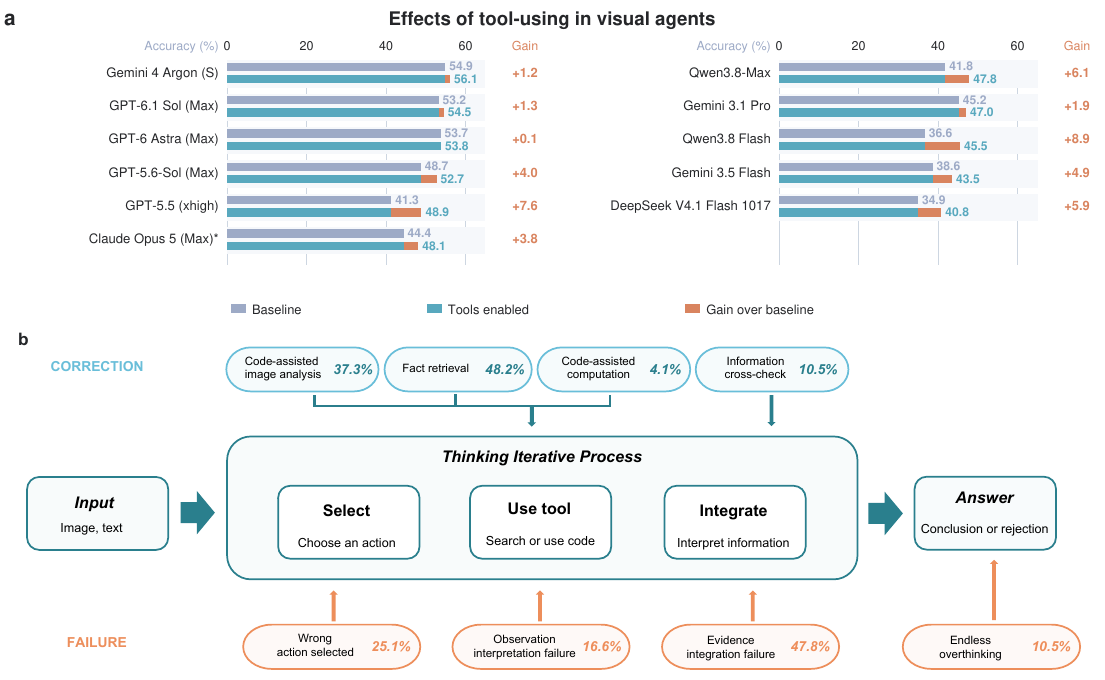}
    \caption{Tool-augmented visual-agent evaluation on \see.
    \textbf{a}, Accuracy with tools and the main leaderboard baseline on the same 1{,}116 questions. Runs may differ in interface. Orange marks the gain in percentage points. S denotes search only; the other models can use search and code. The asterisk marks no-response cases scored as incorrect.
    \textbf{b}, Merged trajectory attribution across eleven models, using the original attribution controls, which can differ from panel a. Corrections arise when models select relevant actions, use tools to inspect or quantify visual evidence, retrieve key external facts, or verify intermediate interpretations. Failures can arise from wrong action selection, misinterpretation of observations, inappropriate integration of retrieved or computed information, or failure to control and terminate the interaction. Percentages use the 706 tool-attributed corrections and 247 regressions that map to the displayed categories, excluding two corrections and three regressions (Supplementary Table~\ref{tab:si-tool-attribution-transitions}).}
    \label{fig:agent_tool_effects}
\end{figure}

Finally, we test whether tool access is sufficient to narrow the capability gap exposed by \see{}. We evaluate eleven MLLMs in a tool-augmented visual-agent setting (Figure~\ref{fig:agent_tool_effects}a). Models receive the same original multimodal inputs as in the standard baseline, but can invoke available tools before producing a final answer. Gemini 4 Argon can use web search; the other ten models can use both web search and a code interpreter. The code interpreter supports programmatic inspection, measurement, and computation over the input image, whereas web search provides external scientific background information. Tool-enabled accuracy exceeds the main leaderboard no-tool baseline in all eleven comparisons. Qwen3.8 Flash shows the largest improvement, increasing from 36.6\% to 45.5\%, while Gemini 4 Argon achieves the highest tool-augmented accuracy of 56.1\%, increasing from 54.9\%. Nevertheless, substantial errors remain, and differences may also reflect evaluation interfaces, indicating that tool access alone does not make MLLMs reliable on \see{}. Detailed configurations are provided in Supplementary Information~\ref{sec:si-tool-augmented-agent-eval}.

Trajectory analysis shows that tool use has bidirectional effects on model outputs (Figure~\ref{fig:agent_tool_effects}b). Across the eleven models, 708 corrections and 250 regressions are substantively attributed to tool use. Figure~\ref{fig:agent_tool_effects}b reports the 706 corrections and 247 regressions mapped to its mechanism categories; the five excluded cases appear in Supplementary Table~\ref{tab:si-tool-attribution-transitions}.

Among the 706 mapped corrections, fact retrieval accounts for 48.2\%, code-assisted image analysis for 37.3\%, information cross-checking for 10.5\%, and code-assisted computation for 4.1\%. These results show that tools can expand the models' ability to inspect visual evidence and access background information, partially compensating for the limitations of static multimodal inference.

Tools can also introduce new errors. Among the 247 mapped regressions, failure modes gather in evidence-integration failures (47.8\%), action-selection failures (25.1\%), observation-interpretation failures (16.6\%), and endless overthinking (10.5\%). Both beneficial and detrimental effects of tools occur in the tool-use trajectory (Figure~\ref{fig:agent_tool_effects}b). This indicates that tool augmentation does not simply provide models with additional information, but turns static multimodal reasoning into an evidence-management problem along the tool-use trajectory. Models must decide which actions are relevant to the current problem, whether tool outputs are reliable, how retrieved or computed information should be integrated with the original experimental observation, and when to stop further interaction.

%% file: sections/05a_error_analysis.tex
\section{Model failures reveal perceptual and inferential weaknesses}

\begin{figure}[!htbp]
    \centering
    \includegraphics[width=0.8\textwidth]{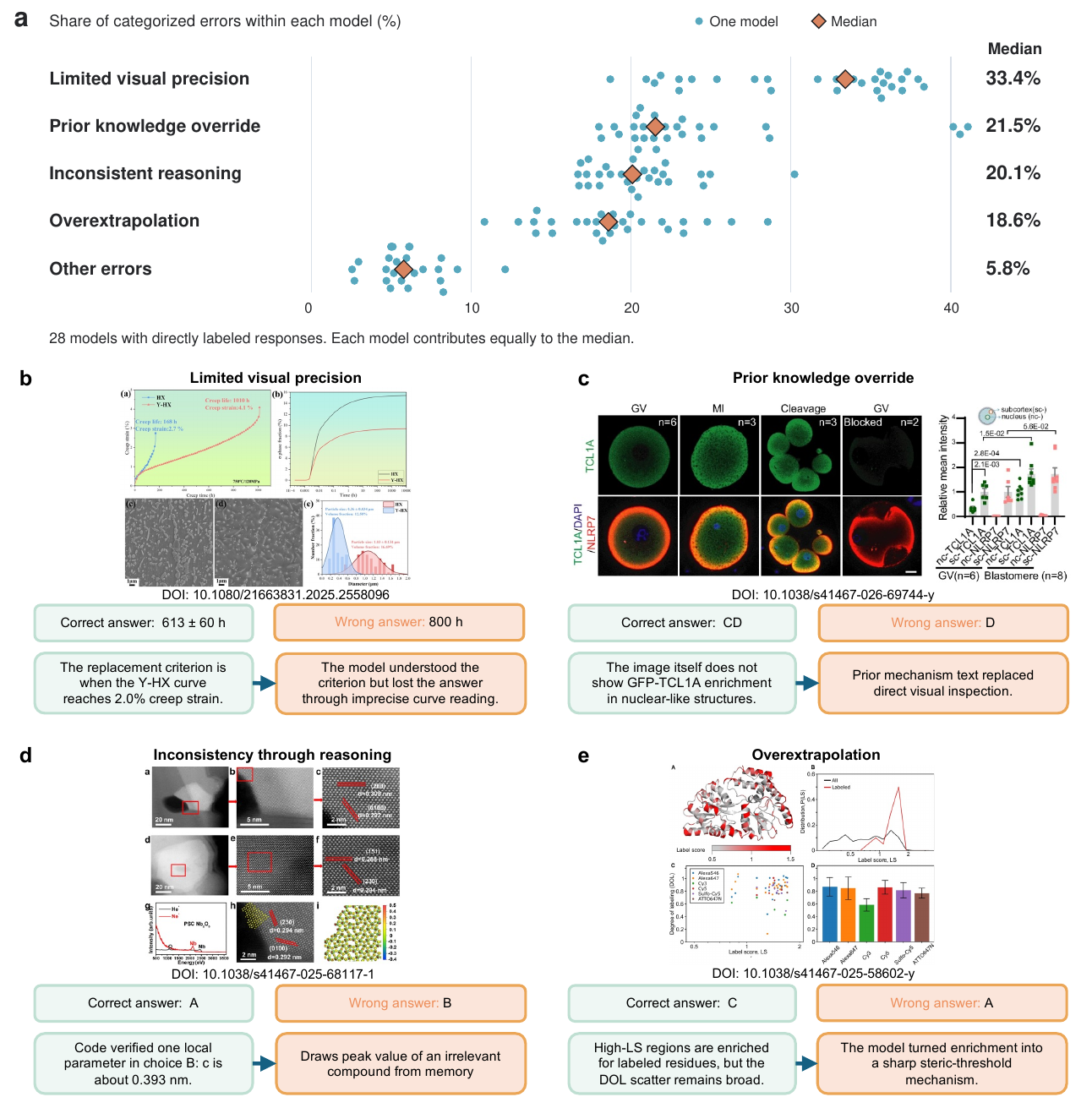}
    \caption{Failure-mode distribution and representative error cases on \see.
    \textbf{a}, Circles show per-model percentages of categorized errors by primary cause; diamonds denote medians across 28 equally weighted models with direct response-level labels. The residual category combines coordination and other failures. Labeling methods, denominators, and expert review are described in Supplementary Information~\ref{sec:si-error-attribution}.
    \textbf{b}, Limited visual precision arises when a model identifies the relevant criterion but cannot accurately read quantitative information from the figure.
    \textbf{c}, Prior knowledge override occurs when a model substitutes a familiar scientific mechanism for direct inspection of the visual evidence.
    \textbf{d}, Inconsistency through reasoning occurs when a model fails to integrate local observations into a coherent global conclusion and instead selects an irrelevant memorized feature.
    \textbf{e}, Overextrapolation occurs when a model turns partial or broad experimental evidence into a stronger mechanistic conclusion than the data support.
    Green boxes indicate the evidence supporting the correct answer, and orange boxes indicate the model's wrong answer and failure pattern.}
    \label{fig:failure_cases}
\end{figure}

To understand why current MLLMs fail on \see, we inspect directly labeled incorrect responses from 28 models and identify four recurring failure modes together with a residual category (Figure~\ref{fig:failure_cases}). The first two are failures in perception, where models do not faithfully extract or use visual evidence, and the next two are failures in inference, where models cannot maintain evidence-constrained reasoning over the available observations, measurements and context. Coordination failures and remaining causes that cannot be categorized are grouped as other errors. Limited visual precision has the largest median share across models (33.4\%), followed by prior knowledge override (21.5\%). The dominant label nevertheless varies by model, indicating that the aggregate distribution conceals heterogeneous failure profiles.

\paragraph{Limited Visual Precision} A substantial proportion of model failures are due to inadequate visual reading precision (tasks requiring the extraction of exact numerical values from spectra, standard curves, or instrument readouts). These failures expose that the visual encoders of current MLLMs are pattern recognizers rather than measurement instruments. They excel at categorical judgments, but perform poorly at fine-grained quantitative extraction. This perceptual inaccuracy can propagate downstream, causing the model to arrive at incorrect conclusions even when it possesses sound domain knowledge and applies valid reasoning logic. For scientific applications where quantitative precision is non-negotiable, this limitation constitutes a fundamental bottleneck that cannot be resolved by improvements to reasoning alone.

\paragraph{Prior Knowledge Override} When visual evidence presented in an image conflicts with patterns prevalent in the training data, models systematically defer to prior knowledge rather than faithfully attending to the actual visual input. Rather than genuinely ``reading'' an image, models tend to infer its content by reverse-mapping from the most frequently encountered concepts in training. This pattern is particularly noticeable in tasks involving molecular structure interpretation typical in biology: models directly match surface-level visual features to high-frequency terms encountered during training instead of decomposing the structure, identifying functional fragments, and progressively deriving the corresponding name or property as a human expert would. This pattern matching shortcut fails when confronted with atypical or non-canonical structures. Current alignment methods produce models that match patterns rather than observe structures. They excel at associating frequent labels with common visuals but lack the bottom-up logic necessary to parse the unknown.

\paragraph{Inconsistency Through Reasoning} A recurring failure mode is the model's inability to maintain a coherent logical framework, often processing question components as disparate units rather than an integrated system. There are two typical patterns. First, models frequently overlook intra-option contradictions. If an option pairs two factually correct but logically incompatible premises, the model tends to evaluate them independently, missing the overall information. Second, there is a clear deficiency in multi-modal integration. When a problem relies on the interplay between several subfigures, models often perform a localized search within one subfigure while neglecting the broader logical constraints established by the remaining figures. This fragmented processing prevents the model from forming a comprehensive relational structure, leading to conclusions that are locally plausible but globally invalid.

\paragraph{Overextrapolation} Another critical failure mode is the tendency to over-infer from partial experimental data. For example, an inquiry requires the synthesis of several independent experiments to reach a conclusion, the current models often treat preliminary or isolated results as conclusive evidence. Models frequently interpolate ``missing'' experimental steps and invent evidence to justify their final claims. This tendency to overextrapolate from individual observations undermines the rigorous evidentiary standards essential to valid scientific inference.

These failure modes indicate that current MLLMs fail in \see not due to a single deficiency, but because perception and inference break down jointly. Visual imprecision, prior-driven interpretation, and weak evidence control often interact, so small errors in reading or selecting evidence cascade into incorrect scientific conclusions. Simply scaling data or model size is therefore unlikely to resolve this gap, because scientific discovery often lies precisely beyond the distribution of previously seen examples.

%% file: sections/05_discussion.tex
\section{Conclusion}
\label{sec:discussion}

We introduced \see to evaluate if current MLLMs can support real laboratory science by reasoning from multimodal experimental evidence. Unlike benchmarks that primarily measure scientific knowledge recall or isolated domain competence, \see is based on peer-reviewed literature and experimental practice in chemistry, biology and materials science. In the standard inference setting, the best-performing model among the 28 MLLMs evaluated reaches only 54.9\% accuracy, and no model exceeds 55\%. This low performance should not be interpreted simply as another difficult benchmark result. Instead, it reveals a mismatch between the abilities captured by many existing scientific benchmarks and the abilities required for the real reasoning process in experimental research.

The additional analyses clarify the nature of this mismatch. Science-specialized models do not close the performance gap. Meanwhile, interdisciplinary analysis shows that \see{} requires models to coordinate concepts, methods, and different types of evidence across disciplinary boundaries. Ablation analysis reveals that when visual evidence is removed, accuracy declines for all 28 models and explicit recognition of missing visual evidence remains limited and highly variable. Together, these findings suggest that the central limitation is not simply what the models know but whether they can extract and organize information from the evidence that is actually available in the problem.

The tool-augmented visual-agent experiment extends this conclusion from static multimodal inference to interactive scientific workflows. Across eleven models, tool access improves accuracy relative to the main leaderboard baselines, with the strongest model reaching 56.1\%, showing that the code interpreter and web search can partially compensate for the limitations of multimodal reasoning. However, trajectory analysis reveals that tool use is not always beneficial. The interaction process can help models inspect visual evidence through code-assisted image analysis, retrieve missing scientific facts, cross-check information, or perform code-assisted computation. Yet it can also introduce new errors through wrong action selection, misinterpretation of tool-derived observations, evidence integration failure, or endless overthinking. The diagnosis of tool-use trajectory provides an explicit view of tool using in scientific reasoning, rather than only the overall accuracy. The main challenge is therefore whether a model can manage tool-derived information by selecting appropriate actions, judging the reliability and relevance of tool outputs, integrating retrieved or computed result with the original information, and stopping when the available evidence is sufficient.

The failure analysis explains why this limitation matters for scientific reasoning. Current MLLMs often miss critical visual evidence, allow prior knowledge to override experimental observations, fail to maintain global logical coherence, and draw conclusions beyond what the evidence supports. These errors are not trivial. These deficiencies highlight a fundamental gap between pattern recognition and the rigorous, evidence-based logic required for authentic scientific inquiry.

These findings suggest a new stage for scientific evaluation where the focus shifts from single-response MLLMs to visual agents and broader agentic scientific systems. Real discovery is rarely completed by answering one question from a fixed input. Instead, it requires an iterative process of proposing hypotheses, selecting tools, transforming or measuring observations, analyzing experimental outputs, and deciding which evidence should be collected next. Future benchmarks for scientific agents should therefore evaluate not only accuracy, but also the quality of the interaction trajectory. This includes assessing whether each action is justified by the available evidence, whether image operations, retrieval, and data analysis are appropriate, whether uncertainty is recognized before the next step, and whether conclusions are revised when new observations contradict prior assumptions. Therefore, agent-level evaluation should extend the evidence-based principle of \see from static experimental reasoning to dynamic scientific decision-making.

Several limitations qualify these conclusions. Final-answer accuracy does not establish reasoning fidelity, and model-group and discipline comparisons are observational rather than causal. The benchmark covers three disciplines and fixed questions rather than complete laboratory workflows. Its 67 withheld items limit full-set reproducibility, so we also report results on the public subset. Tool and no-tool runs may differ in interface, and some paired image-removal runs contain input-version differences; these comparisons therefore support diagnostic rather than strictly causal interpretations.

Overall, \see{} shows that the main barrier to scientifically useful AI is not simply more knowledge, stronger domain specialization, or broader tool access, but the ability to manage multimodal evidence and make justified evidence-bounded inferences from experimental results. The tool-augmented visual-agent analysis shows that this challenge persists in interactive workflows, where tools expand evidence access, but also require reliable action selection, output evaluation, evidence integration, and termination control. The failure analysis reveals the same limitations. Current MLLMs miss critical observations, let prior knowledge override evidence, lose global coherence, and overextend conclusions beyond the data. Future evaluations should therefore test not only what models know, but whether they can transform multimodal experimental observations into new, evidence-supported scientific understanding. This is the missing step toward real scientific discovery.

%% file: sections/07_supplementary.tex

\section{Dataset Details}
\label{sec:si-dataset}

\begin{figure*}[tp]
    \centering
    \includegraphics[width=\textwidth,height=0.78\textheight,keepaspectratio]{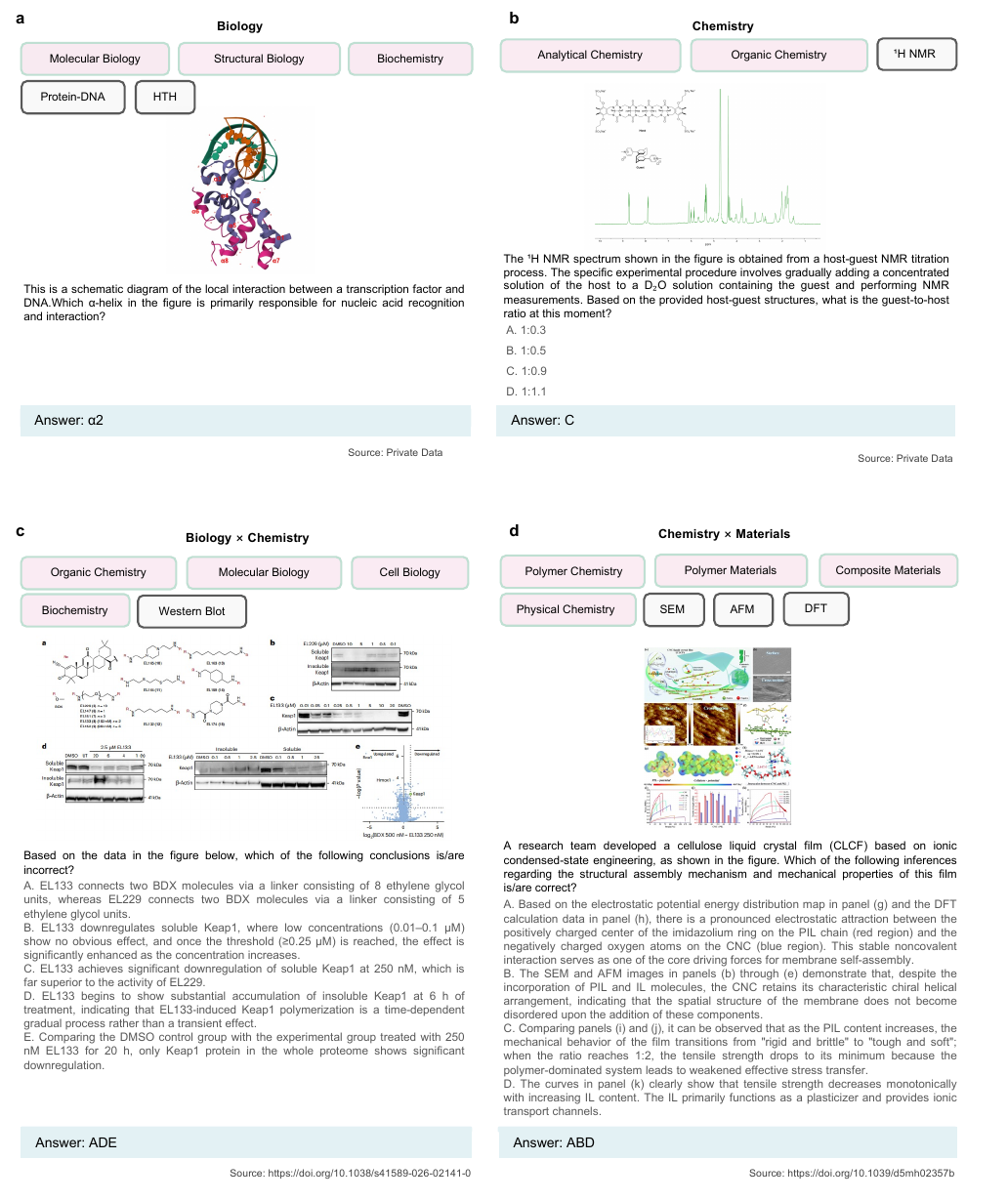}
    \caption{Representative questions in \see. Four examples illustrating the range of disciplinary scopes, evidence types and answer formats in the benchmark.
    \textbf{a}, A single-discipline biology question.
    \textbf{b}, A single-discipline chemistry question.
    \textbf{c}, A cross-disciplinary biology--chemistry question combining chemical structures, Western blots and a volcano plot.
    \textbf{d}, A cross-disciplinary chemistry--materials question that assembles electrostatic potential maps, DFT results, SEM and AFM images and stress--strain curves.
    In every panel, labels with colors annotate sub-fields and outlined labels give the associated experimental technique or knowledge point. Choice answers require exact option-set matches; canonical short answers and numerical answers use the judging rules and tolerances described in Section~\ref{sec:si-eval-prompt}. Panel \textbf{a} requires a canonical short answer, \textbf{b} a single choice, and \textbf{c} and \textbf{d} multiple choices. The source line gives the literature DOI, or ``Private Data'' for questions built on unpublished experimental data contributed by domain experts.}
    \label{fig:data_composition}
\end{figure*}

\see contains 1{,}116 multimodal scientific questions. Each released entry includes the question text, associated visual files, a standard answer, source information, a UUID, discipline labels, and knowledge-point labels (Supplementary Figure~\ref{fig:data_composition}). The associated visual files are entry-level references: they may include images required by the question, but may also include images used in the solution rationale or expert verification, and therefore should not be read as a direct count of question-side figures.

The released question file includes both entries with explicit option lists and entries without a separate option field. The latter include short-answer, numerical, and visually grounded selection tasks whose answers are represented directly as canonical answers. The dataset metadata records three task-type categories: reasoning, measurement, and image processing. The source field is normalized into DOI-like literature sources, personal or experimental sources, and other source text or URL records.

Evidence categories in main-paper Figure~1d are inferred from keywords in the question text and knowledge-point labels. A question contributes equal fractions to all of its matched categories, or to Other when none match; each sub-field column sums to 100\% before rounding.

\subsection{Subject Label Coverage}
\label{sec:si-eval-coverage}
All 1{,}116 questions have at least one of the 17 sub-discipline labels. Tables~\ref{tab:si-label-cardinality} and~\ref{tab:si-discipline-coverage} report the number of labels per question and coverage by main discipline. Table~\ref{tab:si-finegrained} gives the sample size and mean accuracy for each sub-discipline.

\begin{table*}[tp]
\centering
\small
\caption{Distribution of reported subject-label cardinality among questions covered by the 17-label taxonomy. Most questions span multiple sub-disciplines, reflecting the cross-concept, cross-direction, and cross-discipline nature of real experimental scenarios rather than isolated single-topic knowledge.}
\label{tab:si-label-cardinality}
\begin{tabular}{@{}crr@{}}
\toprule
\textbf{Labels per Question} & \textbf{\# Questions} & \textbf{Share (\%)} \\
\midrule
1 &  96 &  8.6 \\
2 & 416 & 37.3 \\
3 & 458 & 41.0 \\
4 & 137 & 12.3 \\
5 &   9 &  0.8 \\
\bottomrule
\end{tabular}
\end{table*}

\begin{table*}[tp]
\centering
\small
\caption{Main-discipline coverage by question count and by label annotation count. Chemistry and biology coverage is high; materials coverage is comparatively lower.}
\label{tab:si-discipline-coverage}
\begin{tabularx}{\textwidth}{@{}>{\raggedright\arraybackslash}Xrrrr@{}}
\toprule
\textbf{Main Discipline} & \textbf{\# Questions} & \textbf{Coverage (\%)} & \makecell{\textbf{\# Label}\\\textbf{Annotations}} & \makecell{\textbf{Share of}\\\textbf{Annotations (\%)}} \\
\midrule
Chemistry & 732 & 65.6 & 1{,}291 & 44.6 \\
Biology   & 538 & 48.2 & 1{,}196 & 41.3 \\
Materials & 363 & 32.5 &   408 & 14.1 \\
\bottomrule
\end{tabularx}
\end{table*}

\paragraph{Cross-Discipline Composition}
Under the 17-label reporting taxonomy, questions labeled with biology only total 370 (33.2\%); questions jointly labeled with chemistry and materials total 346 (31.0\%); chemistry-only questions total 219 (19.6\%); and chemistry-with-biology questions total 164 (14.7\%). Materials-only questions total just 13 (1.2\%). A small number of questions combine all three main disciplines (3 questions, 0.3\%) or biology with materials only (1 question, 0.1\%). This distribution matches the strong overlap between materials science and chemistry in real experimental practice, and accordingly model accuracy on materials labels should be interpreted as reflecting joint chemistry--materials reasoning rather than isolated materials knowledge.

\paragraph{Long-Tail Fine-Grained Labels}
Analytical chemistry (440 questions) is the most frequent reported fine-grained label, followed by biochemistry (344) and physical chemistry (337). At the tail, metallic materials, genetics, immunology, and physiology have only 49, 36, 32, and 19 questions respectively. Accuracy estimates on tail labels are more sensitive to a small number of questions and should be read with sample size in mind.

\section{Evaluation}
\label{sec:si-evaluation}

\subsection{Evaluation Setup}
\label{sec:si-eval-setup}
We evaluate 28 MLLMs, listed in Table~\ref{tab:si-public-subset}. Intern-S1 Pro, Intern-S2, Intern-S2 Preview-35B, and Intern-S2 Preview-397B are science-specialized models. The other 24 models are general-purpose systems. All evaluated models support the multimodal inputs required by the benchmark.

Configurations labeled Max designate extended-thinking runs. The GPT-5.5 row retains its recorded \texttt{xhigh} designation, but its effective effort cannot be verified from the retained records, as detailed below. We do not introduce additional manual prompt optimization, response filtering, or post-hoc correction during evaluation. For questions containing image inputs, images are preprocessed according to the requirements of each model interface, including necessary format conversion, resizing, and input organization. Claude Opus 4.8 (Max) and Claude Opus 5 (Max) have safety-filtered no-response cases on biochemistry questions involving viral biology and pathogen structural characterization; these cases are counted as incorrect under the strict binary scoring protocol.

\see contains 1{,}116 multimodal questions with both explicit-option and open-answer formats. The raw dataset records associated visual files at the entry level; in evaluation, models are given the image inputs retained in the evaluation payload for each question. Model performance is measured by accuracy, where a question is counted as correct only if the final model answer matches the ground-truth answer under the scoring rules described below.

Where both default and Max configurations were evaluated for GPT-6 Astra, GPT-6.1 Sol and Claude Opus 5.5, we report only Max, applying this selection to all three families irrespective of their scores. Gemini 4 Argon, DeepSeek V4.1 Flash, Kimi K3.1 Pro, Grok 4.7, Qwen3.8 Flash, and MiMo V2.6 Pro use their recorded provider-default inference settings. The three refreshed Intern runs also use recorded provider-default settings. Configured API model identifiers and effort settings are recorded. Per-answer backend revision and image-encoding metadata are unavailable for these configurations, limiting verification of the served revision.

Discipline breakdowns cover all 28 models in Tables~\ref{tab:accuracy_discipline} and~\ref{tab:accuracy_discipline_second}. Supplementary Section~\ref{sec:si-eval-textonly} reports the image-removal comparisons, their paired question counts, and the missing-image acknowledgment criterion. Eleven models contribute paired tool-enabled and no-tool scores. The merged trajectory analysis covers eleven models, all using human-in-the-loop attribution with expert review.

\paragraph{Recovered configuration records.}
Available run status and client-initialization logs provide the following configuration details. Request IDs are the gateway-resolved names, rather than the local names with effort suffixes: for example, the Claude and MiMo requests use \modelid{claude-opus-5-5} and \modelid{MiMo-V2.6-Pro}. The corresponding runs use concurrency five and no tools. ``Default'' means that no explicit effort override is recorded; it does not specify a numerical thinking budget or imply that reasoning is disabled. GPT-6.1 Sol (Max) requests \modelid{reasoning.effort=max}; the retained Claude Max implementation enables adaptive thinking with \modelid{output_config.effort=max}. GPT-6 Astra (Max) uses the Responses interface through evamux with request ID \modelid{gpt-6-astra} and \modelid{reasoning.effort=max}; retained response checks confirm both fields. Its leaderboard baseline uses no tools and generation concurrency three.

For leaderboard runs without complete request-level records, temperature, top-$p$, output limits, generation dates, and backend revisions are not fully recoverable. Run names alone do not establish effective inference parameters. In particular, the 461-correct GPT-5.5 leaderboard file is byte-identical to a source artifact named \modelid{data_judged_no_effort_bak.json}; a separate source file contains 477 correct answers. We retain the reported scores and recorded row names, but cannot certify \texttt{xhigh} for the 461-correct responses or equal effort between its no-tool and tool runs.

\paragraph{Image encoding and parameter provenance.}
The retained generation implementation limits the longest image edge to 8,000 pixels, preserving aspect ratio. Images within both the dimension and file-size limits are sent without re-encoding; larger images undergo JPEG compression, with a 2,048-pixel fallback when needed. The retained client configuration assigns file-size thresholds of 3.5 MiB for Claude, 7 for Gemini, 10 for DeepSeek, Kimi and Qwen, and 20 for GPT, Grok and MiMo. These are recovered client implementation settings, not provider-side image resolutions or proof of the bytes sent in every request. The saved image-removal manifests identify this encoder by source hash; some multimodal baseline records lack per-answer encoded-image hashes. We do not substitute present-day provider defaults for missing sampling or token-budget parameters.

\subsection{Prompts and Answer Scoring}
\label{sec:si-eval-prompt}
We use separate prompts for answer generation and answer judging. The generation prompt asks each evaluated model to analyze the question and return a JSON object containing both the analysis and the final answer. The judge compares the candidate response with the reference answer and returns a binary correctness label. Recorded judgments are retained. Saved prompt records identify the Chinese \texttt{default} generation and judging templates and the Chinese question and option fields for the relevant baseline, image-removal, and native-tool runs. Listings~\ref{lst:si-generate-prompt} and~\ref{lst:si-judge-prompt} are English translations of those templates, rather than the literal request language. The recorded prompt-file hash matches the retained Chinese templates.

\begin{lstlisting}[style=promptstyle, caption={English translation of the Chinese default generation prompt.}, label={lst:si-generate-prompt}]
[System]
You are an intelligent assistant. Please read the question and images
carefully, and provide the correct answer.

[User]
Please answer the following question. If it is a choice question, it may
be either single-choice or multiple-choice.

[Question text]:
{question}

[Options]:
{options}

Please first provide the analysis process, and then provide the final
answer. Strictly output the following JSON format, and do not include
Markdown formatting:
{
  "analysis": "your analysis process",
  "answer": "final answer"
}
\end{lstlisting}

\begin{lstlisting}[style=promptstyle, caption={English translation of the Chinese default judging prompt.}, label={lst:si-judge-prompt}]
[System]
You are a strict examiner. Please judge whether the student's answer is
consistent with the reference answer.

[User]
Please judge whether the following answer is correct:

[Question]:
{question}

[Reference answer]:
{reference}

[Student answer] (it may be a short answer, or a full response containing
an analysis process):
{candidate}

[Task]:
1. If the student's answer is a full response containing an analysis
   process, first identify the final answer from it, and then compare it
   with the reference answer.
2. Judge whether the core meaning is consistent. For choice questions,
   the letters must be identical. For fill-in-the-blank or short-answer
   questions, the numerical value or key phrase must be identical. Format
   differences such as "5" and "5.0" are allowed.
3. If the reference answer is a numerical range or contains an error
   tolerance, such as "5-10", "[0.8, 1.2]", "greater than 100",
   "5 +/- 0.5", "about 0.05", or "<3.2", the student's answer is correct
   as long as the numerical value falls within the range. Range
   boundaries are also treated as correct (closed interval). The
   student's answer may also be a range; in this case, judge whether the
   two ranges are substantially consistent, with similar centers and
   widths.
4. If the student refuses to answer or no answer can be identified, mark
   it as incorrect.

Strictly output the following JSON format, and do not include Markdown
formatting:
{
  "correct": true or false,
  "reason": "judgment reason within 50 characters"
}
\end{lstlisting}

For models that support multimodal inputs, images are provided together with the question text in the same evaluation instance. No additional tool use, retrieval, or external browsing is allowed in the standard evaluation setting.

For entries with explicit option lists, including both single-choice and multiple-choice questions, the judge first extracts the final answer from the model response and then checks whether the predicted option letters exactly match the ground-truth option letters. Partial matches are not counted as correct.

For entries without a separate option field, answers are evaluated against canonical short answers or numerical answers. Numerical answers are treated as correct when they match the reference value or fall within the reference range or tolerance specified in the answer. Model outputs that refuse to answer, state that the question cannot be determined, fail to provide a final answer, or otherwise cannot be judged as matching the ground-truth answer are counted as incorrect. Thus, all final evaluation results follow a strict binary scoring protocol: each question is either correct or incorrect.

\subsection{Judge Sensitivity Analysis}
\label{sec:si-judge-sensitivity}

We aligned the Gemini 3.1 Pro multimodal baseline and its GPT-5.5 re-judgment by question UUID. The questions, reference answers, and candidate responses match for all 1{,}116 entries. Both judges returned binary labels for 1{,}111 responses (Table~\ref{tab:si-judge-agreement}). Leaderboard accuracy uses all 1{,}116 questions, counting responses without a binary correctness label as incorrect. This audit covers one model and one input condition.

\begin{table*}[tp]
\centering
\small
\caption{Paired judgments for the Gemini 3.1 Pro multimodal baseline. Agreement is 1,110/1,111 (99.91\%), with Cohen's $\kappa=0.9982$. On all 1,116 questions, Gemini 3.1 Pro scores 504 responses as correct and the independent GPT-5.5 judge scores 505 (+0.09 percentage points). The only disagreement favors the evaluated model under the independent judge.}
\label{tab:si-judge-agreement}
\begin{tabular}{lrr}
\toprule
Gemini 3.1 Pro judgment & GPT-5.5 incorrect & GPT-5.5 correct \\
\midrule
Incorrect & 606 & 1 \\
Correct & 0 & 504 \\
\bottomrule
\end{tabular}
\end{table*}

\subsection{Primary Error Attribution}
\label{sec:si-error-attribution}
\begin{figure*}[tp]
    \centering
    \includegraphics[width=\textwidth]{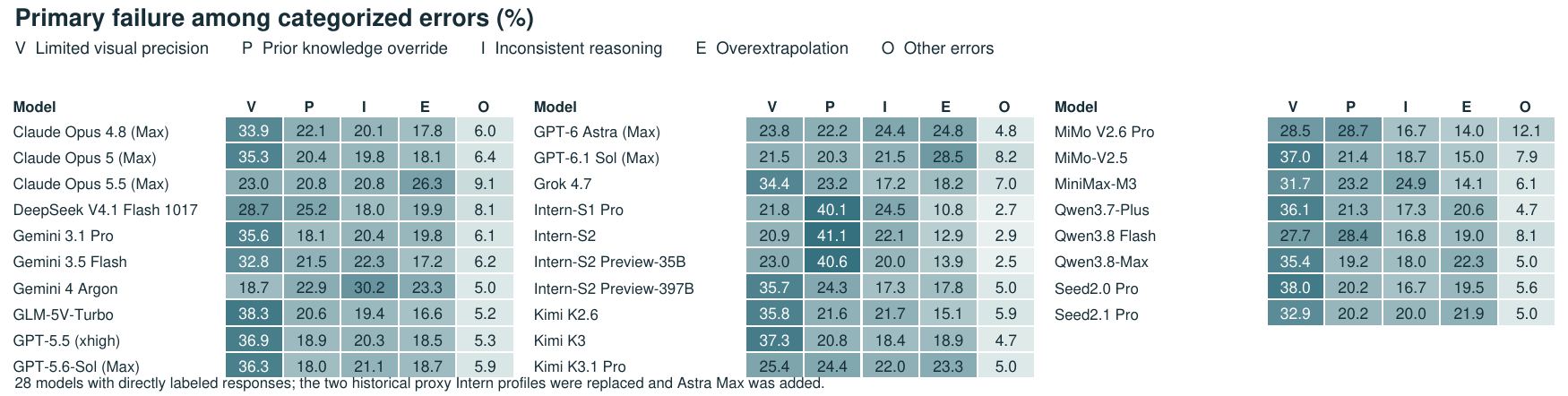}
    \caption{Primary failure modes by model. Each cell gives the percentage of a model's categorized errors assigned to one primary cause. Columns follow the five categories used in main-paper Figure~4a. All 28 rows use direct response-level labels; the two historical proxy Intern profiles have been replaced by new direct measurements. For Claude Opus 5.5 (Max), 274 wrong responses have text for attribution and 500 further errors are empty refusals.}
    \label{fig:si-error-heatmap}
\end{figure*}

Figure~\ref{fig:si-error-heatmap} gives the 28 directly labeled model profiles. The main-paper dot plot includes the same 28 models, and each category's median uses their percentages before rounding.

We apply the same taxonomy to 19,595 labeled incorrect responses across 28 models. GPT-5.6-Luna and Qwen3.8 Flash serve as automated labelers. For each question, the labeler receives the question text, options, reference solution and answer, scoring explanation, and saved model responses. It assigns an ordered list of causes to each response using the textual evidence. The first cause determines the primary label. Domain experts reviewed all 19,595 response-level attributions, including the added Astra labels, to assess whether the assigned primary category matched the error evident in the corresponding response. We combine coordination failures and other residual causes for display.

For Intern-S1 Pro, Intern-S2 Preview-35B, and Intern-S2, we use direct response-level labels for 934, 870, and 750 candidate errors, respectively. Intern-S2 Preview-397B has direct labels for 791 of 792 candidate errors. GPT-6 Astra (Max) contributes 517 directly labeled candidate errors from the completed main-leaderboard baseline used for image removal and SC.

The denominator for each directly labeled model is its number of categorized incorrect responses. These records come from the runs used for attribution. For example, 56 of the 639 GPT-5.5 question UUIDs with error attributions were answered correctly in the leaderboard run.

\subsection{Tool-Augmented Visual-Agent Evaluation Protocol}
\label{sec:si-tool-augmented-agent-eval}
The paired accuracy comparison covers eleven models. Supplementary Table~\ref{tab:si-public-subset-tool} lists these pairs and the Qwen3.8-Max run used for trajectory attribution. Figure~5a compares each tool run with the standard leaderboard baseline; trajectory attribution retains its original controls. All 1,116 questions remain in each accuracy denominator; generation failures, refusals, and non-answers count as incorrect. The following protocol describes the shared configuration of the original six-model subset. To test whether iterative, tool-augmented inference improves performance under more pragmatic research-assistance conditions, we evaluate six models in a visual-agent environment with both web search and a code interpreter: GPT-5.6-Sol (Max), GPT-5.5 (xhigh), Claude Opus 5 (Max), Gemini 3.1 Pro, Qwen3.8-Max, and Gemini 3.5 Flash. For each model, we use the tool capabilities officially provided in its evaluation environment rather than third-party or custom-built substitutes. Each tool-augmented visual-agent run uses the same 1{,}116-question evaluation set as its corresponding standard baseline and receives the same question text, options, and image inputs.

The shared tool configuration is summarised in Table~\ref{tab:si-tool-augmented-agent-config}. The tool-augmented visual-agent setting differs from the standard baseline by allowing each model to form a multi-step interaction trajectory at inference time. At each step, the model may invoke web search to gather external context or use the code interpreter to inspect and manipulate the visual input, perform computation, or verify an intermediate interpretation. The resulting text, numerical output, or processed visual observation is returned to the model and may inform its next action or final answer. No local database, curated retrieval corpus, or domain-specific scientific tool is provided. Tool use is optional rather than forced, and the model may terminate the trajectory and answer directly whenever it judges the available evidence sufficient. This design distinguishes the tool-augmented visual-agent setting from the standard baseline, which requires a final answer from the original multimodal input without external retrieval or code execution.

\paragraph{Native-tool run settings.}
Table~\ref{tab:si-native-request-config} reports request settings for six native-tool runs, including Qwen3.8-Max. All use the Responses interface and concurrency five. The runner supplies no explicit temperature, top-$p$, output-token cap, or maximum tool-call count; these remain provider-controlled rather than a common fixed budget. GPT-6.1 Sol and GPT-6 Astra request \texttt{max} effort; the other four have no explicit effort override. The HTTP client uses a 900-second network timeout per generation attempt (30 seconds for connection establishment), not a uniform 900-second total reasoning budget. A generation pass permits up to three transport attempts; later repair passes preserve successful answers and retry endpoint failures. The code interpreter uses an automatically managed native container; networking is not explicitly blocked by the runner. These settings do not establish an isolated code-only condition. Same-interface no-tool controls are retained separately from the standard leaderboard controls used in the accuracy figure.

\begin{table*}[tp]
\centering\small
\caption{Recovered request settings for six native-tool runs. Search types are the names declared to the gateway. SC also declares \modelid{code_interpreter} with an automatic container; S declares search only. Other sampling and output-budget parameters are omitted by the runner.}
\label{tab:si-native-request-config}
\begin{tabularx}{\textwidth}{@{}>{\raggedright\arraybackslash}Xllll@{}}
\toprule
Model & Gateway & Tools & Search type & Effort override \\
\midrule
GPT-6.1 Sol (Max) & evamux & SC & \modelid{web_search_preview} & \texttt{max} \\
GPT-6 Astra (Max) & evamux & SC & \modelid{web_search_preview} & \texttt{max} \\
Gemini 4 Argon & evamux & S & \modelid{web_search_preview} & None \\
DeepSeek V4.1 Flash 1017 & DashScope & SC & \modelid{web_search} & None \\
Qwen3.8-Max & DashScope & SC & \modelid{web_search} & None \\
Qwen3.8 Flash & DashScope & SC & \modelid{web_search} & None \\
\bottomrule
\end{tabularx}
\end{table*}

\begin{table*}[tp]
\centering
\small
\caption{Tool configuration for the six-model trajectory subset. Each model receives the same benchmark inputs as in the standard multimodal baseline and uses the web-search and code-interpreter capabilities officially provided in its evaluation environment for iterative evidence acquisition and verification during inference. The asterisk on Claude Opus 5 (Max) denotes no-response cases, which are counted as incorrect under the strict binary scoring protocol.}
\label{tab:si-tool-augmented-agent-config}
\begin{tabularx}{\textwidth}{@{}>{\raggedright\arraybackslash}p{3.2cm}X@{}}
\toprule
\textbf{Component} & \textbf{Configuration} \\
\midrule
Evaluated models & Six MLLMs with complete paired standard and tool-augmented visual-agent runs: GPT-5.6-Sol (Max), GPT-5.5 (xhigh), Claude Opus 5 (Max)*, Gemini 3.1 Pro, Qwen3.8-Max, and Gemini 3.5 Flash. \\
Input payload & Same question text, options, and evaluation image inputs as the standard baseline evaluation. \\
Added tools & The web-search and code-interpreter capabilities officially provided in each model's evaluation environment; no third-party or custom-built substitutes are used. No local database, curated retrieval corpus, or specialized scientific tool is enabled. \\
Tool policy & The evaluated model decides whether to call available tools; tool use is optional rather than required for every question. \\
Scoring & Same binary judging protocol as the baseline; stage failures, refusals, unanswered outputs, and outputs without an identifiable final answer are counted as incorrect. \\
\bottomrule
\end{tabularx}
\end{table*}

The tool-augmented visual-agent evaluation contains two stages. In the first stage, the evaluated model analyzes the multimodal question and may construct a variable-length visual-agent trajectory: it selects a tool action, observes the execution result, and either continues gathering or verifying evidence or terminates with the structured answer object described above. In the second stage, only the final answer produced by the trajectory is evaluated, using the same binary correctness protocol as in the baseline experiments. Intermediate tool outputs are retained for trajectory-level diagnostics but are not scored independently.

For multiple-choice questions, the final option letters must match the ground-truth answer exactly. For short-answer and numerical questions, the answer is compared against the standardized canonical answer and any expert-defined tolerance. Tool-augmented execution failures, refusals, unanswered outputs, and outputs from which no final answer can be identified are counted as incorrect, matching the conservative treatment used in the standard evaluation.

\paragraph{Human-in-the-loop trajectory attribution.}
Across all eleven models, we compare each paired standard and tool-augmented visual-agent trajectory for which correctness changes between the two settings. We use a human-in-the-loop procedure in which an attribution model first performs a structured preliminary analysis using the question, reference answer, standard response, tool-augmented visual-agent response, and available tool-call trace. It identifies the key step associated with the improvement or deterioration, determines whether and how web search or code execution plays a substantive role, and proposes a mechanism category with supporting evidence. Domain experts then inspect the original responses and tool trace, verify the proposed attribution against the task-specific experimental context, and revise the category or rationale when necessary. Improvement mechanisms include code-assisted figure analysis, retrieval of key scientific facts, search-based confirmation, and code-based computation. Tool-related detrimental mechanisms are organized into four recurring failure categories: action-selection failure, observation-interpretation failure, evidence-integration failure, and execution-control failure (reported as ``endless overthinking'' in the main-paper figure). Their relationship to the fine-grained attribution types is summarized in Table~\ref{tab:si-tool-failure-taxonomy}. The resulting expert-validated attribution is used as a diagnostic analysis of the interaction trajectory rather than as an additional correctness score.

\begin{table*}[tp]
\centering
\small
\caption{Mapping between the recurring categories of tool-related failure and their fine-grained manifestations in the trajectory attribution analysis. Counts assign one primary category to each of the 247 mapped tool-attributed regressions across eleven models.}
\label{tab:si-tool-failure-taxonomy}
\begin{tabularx}{\textwidth}{@{}>{\raggedright\arraybackslash}p{4.0cm}X@{}}
\toprule
\textbf{Failure category} & \textbf{Fine-grained manifestations} \\
\midrule
Action-selection failure (62/247) & Inappropriate use of code, including an unsuitable image-processing method, data-processing assumption, calculation target, or retrieval direction. \\
Observation-interpretation failure (41/247) & Code-logic or computation error; tool-mediated misreading of figures, crops, measurements, or computed outputs. \\
Evidence-integration failure (118/247) & Over-reliance on retrieved information; noisy or misleading retrieval; inappropriate weighting of tool-derived evidence relative to the question-specific experimental evidence. \\
Execution-control failure (26/247) & Excessive tool use and context degradation; unproductive repetition; failure to stop or produce a final answer. \\
\bottomrule
\end{tabularx}
\end{table*}

Across eleven models, 1,831 paired instances change correctness: 1,150 corrections and 681 regressions, yielding a net gain of 469 correct predictions. Trajectory attribution identifies tool contributions in 708 corrections and 250 regressions (Table~\ref{tab:si-tool-attribution-transitions}). For Figure~5b, we exclude one source-data-extraction correction, one direct-answer-acquisition correction, and three missing-final-answer regressions whose traces do not establish endless overthinking. The mapped correction denominator is 706, comprising 340 fact-retrieval, 263 figure-analysis, 74 confirmation, and 29 computation cases. The mapped regression denominator is 247, comprising 118 evidence-integration, 62 action-selection, 41 observation-interpretation, and 26 endless-overthinking cases. The observation-interpretation group includes one custom tool-evidence-misinterpretation case. The other five models use GPT-6.1 Sol for structured preliminary attribution, followed by the same expert review. Qwen3.8-Max contributes its attribution trajectories, while the accuracy panel compares its paired tool run with the standard leaderboard baseline. Astra attribution uses all 1,116 main-leaderboard baseline pairs without exclusions. Other attribution controls can also differ from those in the accuracy panel.

\begin{table*}[tp]
\centering
\small
\caption{Paired correctness transitions and trajectory attribution by model. Wrong$\rightarrow$Correct and Correct$\rightarrow$Wrong report answer changes in the original attribution pairs, whose controls can differ from the leaderboard baselines in Figure~5a. Tool-attributed improvements and regressions report the subsets for which web search or code execution is identified as playing a substantive role. Percentages are calculated relative to the corresponding number of improvements or regressions for each model. Counts cover eleven models and include the five cases excluded from the category percentages in Figure~5b: two corrections and three regressions. Qwen3.8-Max uses its original attribution run; Astra uses the completed main leaderboard baseline for attribution.}
\label{tab:si-tool-attribution-transitions}
\begin{tabularx}{\textwidth}{@{}>{\raggedright\arraybackslash}Xccccc@{}}
\toprule
\textbf{Model} & \makecell{\textbf{Wrong$\rightarrow$}\\\textbf{Correct}} & \makecell{\textbf{Correct$\rightarrow$}\\\textbf{Wrong}} & \makecell{\textbf{Net}\\\textbf{change}} & \makecell{\textbf{Tool-attributed}\\\textbf{improvements}} & \makecell{\textbf{Tool-attributed}\\\textbf{regressions}} \\
\midrule
Claude Opus 5 (Max) & 80 & 38 & +42 & 24 (30.0\%) & 8 (21.1\%) \\
Gemini 3.1 Pro & 95 & 74 & +21 & 24 (25.3\%) & 15 (20.3\%) \\
Gemini 3.5 Flash & 132 & 77 & +55 & 75 (56.8\%) & 28 (36.4\%) \\
GPT-5.5 (xhigh) & 127 & 42 & +85 & 103 (81.1\%) & 19 (45.2\%) \\
GPT-5.6-Sol (Max) & 113 & 68 & +45 & 101 (89.4\%) & 50 (73.5\%) \\
Qwen3.8-Max & 101 & 43 & +58 & 84 (83.2\%) & 25 (58.1\%) \\
\midrule
GPT-6.1 Sol (Max) & 64 & 50 & +14 & 48 (75.0\%) & 21 (42.0\%) \\
GPT-6 Astra (Max) & 53 & 52 & +1 & 42 (79.2\%) & 20 (38.5\%) \\
Gemini 4 Argon & 58 & 45 & +13 & 4 (6.9\%) & 1 (2.2\%) \\
DeepSeek V4.1 Flash 1017 & 153 & 97 & +56 & 81 (52.9\%) & 21 (21.6\%) \\
Qwen3.8 Flash & 174 & 95 & +79 & 122 (70.1\%) & 42 (44.2\%) \\
\midrule
\textbf{Overall} & \textbf{1,150} & \textbf{681} & \textbf{+469} & \textbf{708 (61.6\%)} & \textbf{250 (36.7\%)} \\
\bottomrule
\end{tabularx}
\end{table*}

\subsection{Public-Subset Reproducibility}
\label{sec:si-public-subset}

The main evaluation includes all 1{,}116 questions, including 67 withheld because they contain unpublished experimental data. To support reproducibility, we also compute accuracy on the 1{,}049 released questions identified by the \modelid{open_source} field. Public-subset model responses and extracted final answers are released alongside the correctness labels. The public and full-set no-tool scores closely track one another across models (Table~\ref{tab:si-public-subset}). Table~\ref{tab:si-public-subset} reports standard scores for all 28 models. Table~\ref{tab:si-public-subset-tool} gives paired tool comparisons for eleven models. The no-tool scores use the standard leaderboard baselines for all models. Tool and baseline runs may differ in interface. The trajectory-attribution analysis retains its original control responses, which differ from the leaderboard runs for some models; its transition counts should therefore not be inferred from this table.

\begin{table*}[tp]
\centering
\small
\caption{Full-set and public-subset accuracy under the standard no-tool evaluation setting. The public subset contains the 1{,}049 released questions and the full set contains all 1{,}116 questions. Delta is public minus full accuracy in percentage points, computed before rounding. Asterisks mark Claude runs with refusals or safety-filtered non-responses counted as incorrect, including 500 refusals for Opus 5.5 (Max).}
\label{tab:si-public-subset}
\begin{tabularx}{\textwidth}{@{}>{\raggedright\arraybackslash}Xrrrrr@{}}
\toprule
\textbf{Model} & \makecell{\textbf{Full}\\\textbf{Correct}} & \makecell{\textbf{Full}\\\textbf{Acc. (\%)}} & \makecell{\textbf{Public}\\\textbf{Correct}} & \makecell{\textbf{Public}\\\textbf{Acc. (\%)}} & \makecell{\textbf{Delta}\\\textbf{(pp)}} \\
\midrule
Gemini 4 Argon & 613/1{,}116 & 54.9 & 571/1{,}049 & 54.4 & -0.5 \\
GPT-6 Astra (Max) & 599/1{,}116 & 53.7 & 555/1{,}049 & 52.9 & -0.8 \\
GPT-6.1 Sol (Max) & 594/1{,}116 & 53.2 & 549/1{,}049 & 52.3 & -0.9 \\
GPT-5.6-Sol (Max)    & 543/1{,}116 & 48.7 & 503/1{,}049 & 48.0 & -0.7 \\
Gemini 3.1 Pro       & 504/1{,}116 & 45.2 & 478/1{,}049 & 45.6 & +0.4 \\
Claude Opus 5 (Max)* & 495/1{,}116 & 44.4 & 465/1{,}049 & 44.3 & -0.0 \\
Kimi K3.1 Pro & 493/1{,}116 & 44.2 & 457/1{,}049 & 43.6 & -0.6 \\
Qwen3.8-Max          & 466/1{,}116 & 41.8 & 439/1{,}049 & 41.8 & +0.1 \\
GPT-5.5 (xhigh)        & 461/1{,}116 & 41.3 & 433/1{,}049 & 41.3 & -0.0 \\
Kimi K3              & 432/1{,}116 & 38.7 & 403/1{,}049 & 38.4 & -0.3 \\
Gemini 3.5 Flash     & 431/1{,}116 & 38.6 & 410/1{,}049 & 39.1 & +0.5 \\
Grok 4.7 & 413/1{,}116 & 37.0 & 391/1{,}049 & 37.3 & +0.3 \\
Qwen3.8 Flash & 409/1{,}116 & 36.6 & 383/1{,}049 & 36.5 & -0.1 \\
Claude Opus 4.8 (Max)* & 406/1{,}116 & 36.4 & 377/1{,}049 & 35.9 & -0.4 \\
DeepSeek V4.1 Flash 1017 & 389/1{,}116 & 34.9 & 361/1{,}049 & 34.4 & -0.4 \\
Seed2.1 Pro          & 377/1{,}116 & 33.8 & 350/1{,}049 & 33.4 & -0.4 \\
Intern-S2            & 366/1{,}116 & 32.8 & 345/1{,}049 & 32.9 & +0.1 \\
Qwen3.7-Plus         & 365/1{,}116 & 32.7 & 346/1{,}049 & 33.0 & +0.3 \\
Claude Opus 5.5 (Max)* & 342/1{,}116 & 30.6 & 329/1{,}049 & 31.4 & +0.7 \\
Intern-S2 Preview-397B & 324/1{,}116 & 29.0 & 302/1{,}049 & 28.8 & -0.2 \\
Seed2.0 Pro         & 315/1{,}116 & 28.2 & 298/1{,}049 & 28.4 & +0.2 \\
MiniMax-M3           & 314/1{,}116 & 28.1 & 302/1{,}049 & 28.8 & +0.7 \\
Kimi K2.6            & 286/1{,}116 & 25.6 & 266/1{,}049 & 25.4 & -0.3 \\
MiMo V2.6 Pro & 282/1{,}116 & 25.3 & 269/1{,}049 & 25.6 & +0.4 \\
GLM-5V-Turbo         & 259/1{,}116 & 23.2 & 245/1{,}049 & 23.4 & +0.1 \\
Intern-S2 Preview-35B    & 246/1{,}116 & 22.0 & 234/1{,}049 & 22.3 & +0.3 \\
MiMo-V2.5            & 196/1{,}116 & 17.6 & 185/1{,}049 & 17.6 & +0.1 \\
Intern-S1 Pro        & 182/1{,}116 & 16.3 & 173/1{,}049 & 16.5 & +0.2 \\
\bottomrule
\end{tabularx}
\end{table*}

\begin{table*}[tp]
\centering
\small
\caption{Paired no-tool and tool-enabled scores. Cells give correct answers and accuracy (\%) on the full set (1,116) or public subset (1,049). S enables search and SC enables search and code. The asterisk marks safety-filtered non-responses counted as incorrect. Controls are the main leaderboard baselines for all models; interfaces can differ between runs.}
\label{tab:si-public-subset-tool}
\begin{tabularx}{\textwidth}{@{}Xcrrrr@{}}
\toprule
Model & Tools & Full control & Full tools & Public control & Public tools \\
\midrule
\multicolumn{6}{l}{\emph{Pairs used for trajectory attribution}} \\
GPT-5.6-Sol (Max) & SC & 543 (48.7) & 588 (52.7) & 503 (48.0) & 547 (52.1) \\
GPT-5.5 (xhigh) & SC & 461 (41.3) & 546 (48.9) & 433 (41.3) & 511 (48.7) \\
Claude Opus 5 (Max)* & SC & 495 (44.4) & 537 (48.1) & 465 (44.3) & 502 (47.9) \\
Gemini 3.1 Pro & SC & 504 (45.2) & 525 (47.0) & 478 (45.6) & 497 (47.4) \\
Qwen3.8-Max (attribution run) & SC & 466 (41.8) & 524 (47.0) & 439 (41.8) & 488 (46.5) \\
Gemini 3.5 Flash & SC & 431 (38.6) & 486 (43.5) & 410 (39.1) & 459 (43.8) \\
\midrule
\multicolumn{6}{l}{\emph{Additional native-tool runs}} \\
GPT-6 Astra (Max) & SC & 599 (53.7) & 600 (53.8) & 555 (52.9) & 557 (53.1) \\
GPT-6.1 Sol (Max) & SC & 594 (53.2) & 608 (54.5) & 549 (52.3) & 566 (54.0) \\
Gemini 4 Argon & S & 613 (54.9) & 626 (56.1) & 571 (54.4) & 582 (55.5) \\
Qwen3.8-Max & SC & 466 (41.8) & 534 (47.8) & 439 (41.8) & 496 (47.3) \\
Qwen3.8 Flash & SC & 409 (36.6) & 508 (45.5) & 383 (36.5) & 474 (45.2) \\
DeepSeek V4.1 Flash 1017 & SC & 389 (34.9) & 455 (40.8) & 361 (34.4) & 419 (39.9) \\
\bottomrule
\end{tabularx}
\end{table*}

\input{sections/optional/text_only_ablation_supplementary}

\paragraph{Qwen3.8 Flash tool attribution.}
The paired runs yield 174 corrections and 95 regressions. Preliminary trajectory attribution with GPT-6.1 Sol assigns a substantive tool role to 122/174 corrections (70.1\%) and 42/95 regressions (44.2\%). All 269 changed cases have a parsed preliminary result. One case required a non-streaming retry with the same model and prompt. Domain experts subsequently review and, where necessary, revise these attributions, which contribute to the merged eleven-model analysis.

Among the 122 tool-attributed corrections, 75 involve retrieval of key facts, 22 code-assisted figure analysis, 17 search confirmation, seven code-based computation, and one direct-answer-acquisition case excluded from the displayed categories. Among the 42 tool-attributed regressions, 23 involve excessive reliance on search results, 13 inappropriate code use, and two each search noise, figure misinterpretation, and misleading search results. The remaining 52 corrections and 53 regressions receive no substantive tool attribution.

\paragraph{GPT-6 Astra tool attribution.}
The main-leaderboard baseline and SC runs yield 53 corrections and 52 regressions across all 1,116 pairs without exclusions. Preliminary GPT-6.1 Sol attribution assigns a substantive search or code role to 42 corrections and 20 regressions, all of which map to the displayed categories; the other 43 changed cases have no substantive tool role and are not counted in the tool-attributed totals. All 105 changed cases have parsed preliminary results: 103 prior attributions were reused after exact case-input checks, and two newly exposed transitions were attributed with the same model and prompt. Domain experts subsequently review and, where necessary, revise these attributions. Together with Qwen3.8 Flash, all 374 changed cases in these two added analyses have parsed preliminary results and expert-reviewed final attributions.

\subsection{Per-Discipline Results}
\label{sec:si-eval-finegrained}
\input{tables/accuracy_by_subdiscipline_model.tex}
\input{tables/accuracy_by_subdiscipline_model_second.tex}

We report performance for all 28 leaderboard models across three main disciplines (chemistry, biology, materials) and the final set of 17 fine-grained discipline labels. Labels are taken from the frozen reporting taxonomy and joined to each model's original correctness labels by question UUID. Original questions, reference answers, and judgments remain unchanged. Because \see uses multi-label annotation, a question may contribute to multiple labels; each question contributes once within each main discipline. Means weight models equally.

Tables~\ref{tab:accuracy_discipline} and~\ref{tab:accuracy_discipline_second} give the 28 models' sub-discipline profiles in two groups of 14, ordered by overall accuracy. Bold identifies the best score across all 28 models for each label. Table~\ref{tab:si-finegrained} gives question counts and mean accuracy. Cross-main-discipline questions carry labels from at least two main disciplines, including questions spanning all three; label-count comparisons use the number of distinct reported sub-discipline labels.

\begin{table*}[tp]
\centering
\small
\caption{Mean accuracy across all 28 leaderboard models for each reported fine-grained discipline label, weighting models equally. Labels with smaller sample sizes (e.g., Immunology, Physiology, Genetics) should be interpreted with care due to their limited support.}
\label{tab:si-finegrained}
\begin{tabularx}{\textwidth}{@{}>{\raggedright\arraybackslash}Xlrr@{}}
\toprule
\textbf{Sub-discipline} & \textbf{Main Discipline} & \textbf{\# Questions} & \textbf{Mean Acc. (\%)} \\
\midrule
Organic Chemistry                & Chemistry  & 230 & 41.6 \\
Analytical Chemistry             & Chemistry  & 440 & 38.1 \\
Metallic Materials               & Materials  & 49  & 37.6 \\
Genetics                         & Biology    & 36  & 37.0 \\
Cell Biology                     & Biology    & 182 & 36.6 \\
Immunology                       & Biology    & 32  & 36.4 \\
Molecular Biology                & Biology    & 261 & 34.8 \\
Biochemistry                     & Biology    & 344 & 34.4 \\
Inorganic Non-metallic Materials & Materials  & 103 & 33.9 \\
Biophysics                       & Biology    & 156 & 33.5 \\
Structural Biology               & Biology    & 166 & 33.0 \\
Physiology                       & Biology    & 19  & 32.0 \\
Physical Chemistry               & Chemistry  & 337 & 31.8 \\
Composite Materials              & Materials  & 68  & 31.3 \\
Polymer Chemistry and Physics    & Chemistry  & 174 & 31.3 \\
Inorganic Chemistry              & Chemistry  & 110 & 30.5 \\
Organic and Polymer Materials    & Materials  & 188 & 27.8 \\
\bottomrule
\end{tabularx}
\end{table*}

\input{tables/interdisciplinary_28}

%% file: sections/optional/text_only_ablation_supplementary.tex
\subsection{Text-Only Ablation}
\label{sec:si-eval-textonly}
To examine whether models recognize missing visual evidence, we compare with-image and text-only runs on all 1,116 question UUIDs for each model. We report two complementary diagnostics: final-answer accuracy without images and explicit acknowledgment of missing visual evidence under the original answer-generation prompt.

\begin{figure*}[tp]
    \centering
    \includegraphics[width=0.96\textwidth]{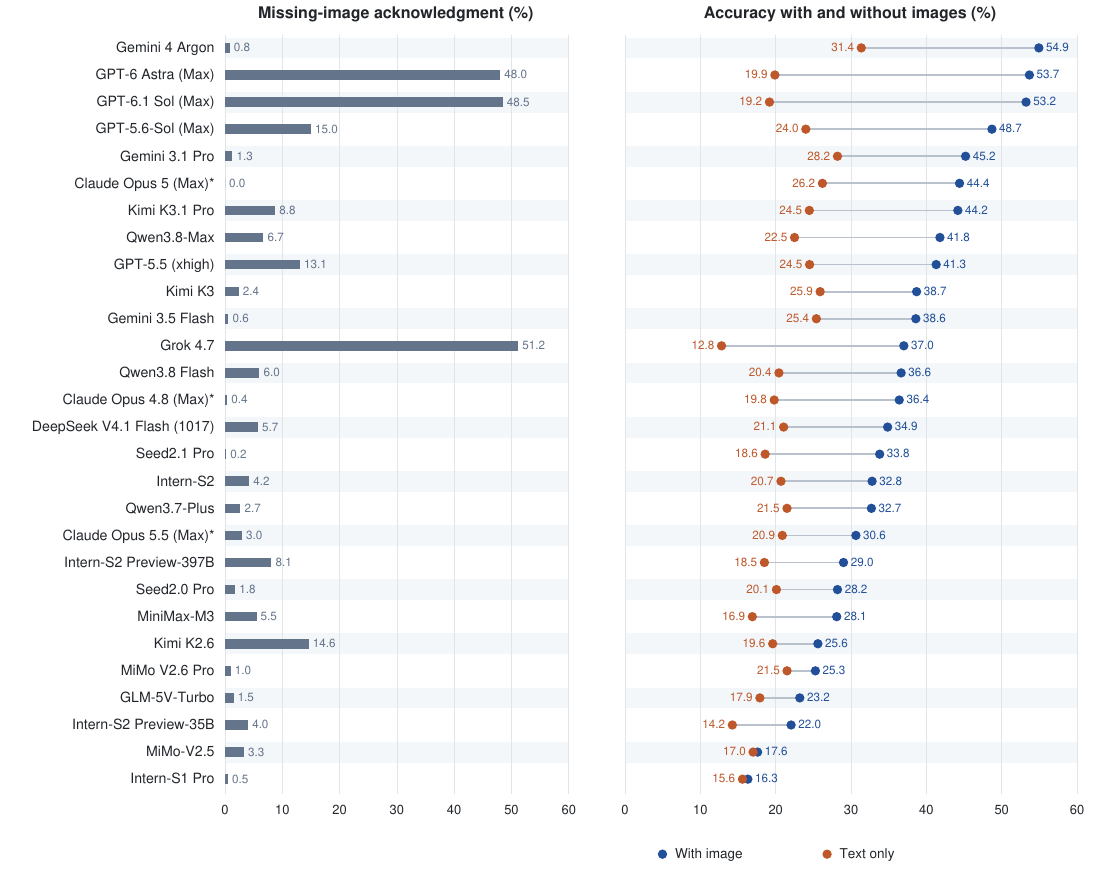}
    \caption{Image removal and missing-image acknowledgment across 28 models. Blue and orange dots show accuracy with and without images on the same 1,116 question UUIDs within each model. Gray bars show the percentage of text-only responses that cite missing visual evidence and withhold a concrete answer. Models are ordered by with-image accuracy. Asterisks mark safety-filtered non-answers scored as incorrect.}
    \label{fig:si-full-set-image-pairs}
\end{figure*}

Figure~3 in the main paper shows paired accuracies for 28 models, and Figure~\ref{fig:si-full-set-image-pairs} additionally shows missing-image acknowledgment rates. Table~\ref{tab:si-added-image-pairs} reports full-set and public-subset results and missing-image acknowledgment rates. Accuracy uses all 1,116 question UUIDs in both conditions and counts refusals, non-answers, and generation errors as incorrect. Across all 28 models, the multimodal runs have 11,102 correct answers and the text-only runs have 6,570 across 31,248 instances per condition, a mean gap of 14.5 percentage points. The spread across models narrows from 38.6 to 18.5 points.

For GPT-5.6-Sol, Claude Opus 5, Qwen3.8-Max, Kimi K3, Seed2.1 Pro, and Intern-S2 Preview-397B, the recorded question, options, and reference fields match across conditions. Their multimodal runs give 1,123 more correct answers over 6,696 pairs (16.8 percentage points). The models in Table~\ref{tab:si-added-image-pairs} also have matched question text, options, image references, and reference answers, as detailed below. For the ten remaining models, input-version differences include option formatting and explicit numerical tolerances. UUID alignment establishes a common question roster, but the 28-model aggregate gap also reflects input-version differences.

Across all 28 text-only runs, 2,888 out of 31{,}248 instances (9.2\%) are classified as explicit acknowledgments that necessary image evidence is missing. This measures expressed recognition of missing images, rather than calibrated uncertainty or the factual support of every generated answer.

In this ablation setting, only explicit missing-image acknowledgments contribute to the acknowledgment rate. Empty outputs, generation errors, generic refusals, and other non-answer cases are not counted toward this metric because they do not directly show recognition that visual evidence is unavailable. For Table~\ref{tab:si-textonly-missing-image}, the denominator includes all 31{,}248 text-only instances. Lack of acknowledgment alone does not establish hallucination: responses may use textual cues, and the generation prompt requests an answer rather than a calibrated abstention decision. Accuracy and acknowledgment rates are therefore interpreted separately.

\input{tables/text_only_missing_image.tex}

\paragraph{Image removal under fixed inputs.}
For each model in Table~\ref{tab:si-added-image-pairs}, we matched the two conditions by question UUID and checked the question text, options, image references, and reference answer. Both conditions use the same prompt and inference settings without tools. All 1,116 questions remain in both accuracy denominators. Explicit refusals, budget terminations, generation failures, and non-answers count as incorrect. Public-subset scores use the same 1,049 released questions in both conditions. An automated classifier applies the missing-image rubric to the text-only responses of each model in the table. The classifier receives the first 2,000 characters of the question, the first 4,000 characters of the response, and the extracted final answer. A response counts when it explicitly cites the missing image and withholds a concrete answer. A guess accompanied by a missing-image caveat counts as an attempted answer. Empty responses remain in the acknowledgment denominator and do not count as acknowledgments.

\begin{table*}[tb]
\centering\small
\caption{Full-set and public-subset accuracy (\%) with and without images. All models use the same 1,116 question UUIDs for the benchmark and 1,049 released UUIDs for the public subset. Ack. is the percentage of all text-only responses classified as withholding an answer because the image is missing. The asterisk marks safety-filtered non-answers counted as incorrect.}
\label{tab:si-added-image-pairs}
\begin{tabular}{lrrrrr}
\toprule
 & \multicolumn{3}{c}{Benchmark (1,116 questions)} & \multicolumn{2}{c}{Public subset (1,049 questions)}\\
\cmidrule(lr){2-4}\cmidrule(lr){5-6}
Model & With image & Text only & Ack. & With image & Text only\\
\midrule
Gemini 4 Argon & 54.9 & 31.4 & 0.8 & 54.4 & 31.6\\
GPT-6 Astra (Max) & 53.7 & 19.9 & 48.0 & 52.9 & 19.4\\
GPT-6.1 Sol (Max) & 53.2 & 19.2 & 48.5 & 52.3 & 18.9\\
Kimi K3.1 Pro & 44.2 & 24.5 & 8.8 & 43.6 & 24.1\\
Grok 4.7 & 37.0 & 12.8 & 51.2 & 37.3 & 13.2\\
Qwen3.8 Flash & 36.6 & 20.4 & 6.0 & 36.5 & 20.0\\
DeepSeek V4.1 Flash (1017) & 34.9 & 21.1 & 5.7 & 34.4 & 20.9\\
Intern-S2 & 32.8 & 20.7 & 4.2 & 32.9 & 20.5\\
Claude Opus 5.5 (Max)* & 30.6 & 20.9 & 3.0 & 31.4 & 21.0\\
MiMo V2.6 Pro & 25.3 & 21.5 & 1.0 & 25.6 & 21.7\\
Intern-S2 Preview-35B & 22.0 & 14.2 & 4.0 & 22.3 & 14.6\\
Intern-S1 Pro & 16.3 & 15.6 & 0.5 & 16.5 & 15.9\\
\bottomrule
\end{tabular}
\end{table*}

%% file: tables/text_only_missing_image.tex
\begin{table*}[tp]
\centering
\small
\caption{Explicit missing-image acknowledgments across 28 models. The denominator is all text-only instances for each model; the numerator includes only responses with \modelid{refusal\_type=no\_image}, i.e., responses explicitly classified as acknowledging missing image or visual evidence. Empty outputs, generation errors, generic refusals, and other non-answer cases are not counted.}
\label{tab:si-textonly-missing-image}
\begin{tabularx}{\textwidth}{@{}>{\raggedright\arraybackslash}Xrrr@{}}
\toprule
\textbf{Model} & \textbf{Total} & \makecell{\textbf{Explicit Missing-}\\\textbf{Image Ack.}} & \textbf{Rate (\%)} \\
\midrule
Grok 4.7 & 1{,}116 & 571 & 51.2 \\
GPT-6.1 Sol (Max) & 1{,}116 & 541 & 48.5 \\
GPT-6 Astra (Max) & 1{,}116 & 536 & 48.0 \\
GPT-5.6-Sol (Max) & 1{,}116 & 167 & 15.0 \\
Kimi K2.6 & 1{,}116 & 163 & 14.6 \\
GPT-5.5 (xhigh) & 1{,}116 & 146 & 13.1 \\
Kimi K3.1 Pro & 1{,}116 & 98 & 8.8 \\
Intern-S2 Preview-397B & 1{,}116 & 90 & 8.1 \\
Qwen3.8-Max & 1{,}116 & 75 & 6.7 \\
Qwen3.8 Flash & 1{,}116 & 67 & 6.0 \\
DeepSeek V4.1 Flash (1017) & 1{,}116 & 64 & 5.7 \\
MiniMax-M3 & 1{,}116 & 61 & 5.5 \\
Intern-S2 & 1{,}116 & 47 & 4.2 \\
Intern-S2 Preview-35B & 1{,}116 & 45 & 4.0 \\
MiMo-V2.5 & 1{,}116 & 37 & 3.3 \\
Claude Opus 5.5 (Max)* & 1{,}116 & 33 & 3.0 \\
Qwen3.7-Plus & 1{,}116 & 30 & 2.7 \\
Kimi K3 & 1{,}116 & 27 & 2.4 \\
Seed2.0 Pro & 1{,}116 & 20 & 1.8 \\
GLM-5V-Turbo & 1{,}116 & 17 & 1.5 \\
Gemini 3.1 Pro & 1{,}116 & 14 & 1.3 \\
MiMo V2.6 Pro & 1{,}116 & 11 & 1.0 \\
Gemini 4 Argon & 1{,}116 & 9 & 0.8 \\
Gemini 3.5 Flash & 1{,}116 & 7 & 0.6 \\
Intern-S1 Pro & 1{,}116 & 6 & 0.5 \\
Claude Opus 4.8 (Max)* & 1{,}116 & 4 & 0.4 \\
Seed2.1 Pro & 1{,}116 & 2 & 0.2 \\
Claude Opus 5 (Max)* & 1{,}116 & 0 & 0.0 \\
\midrule
\textbf{Total / Mean} & 31{,}248 & 2{,}888 & 9.2 \\
\bottomrule
\end{tabularx}
\vspace{2pt}
{\footnotesize\raggedright \textsuperscript{*}Safety-filtered no-response cases, counted as incorrect under the strict binary scoring protocol.\par}
\end{table*}

%% file: tables/accuracy_by_subdiscipline_model.tex
\begin{table}[t]
\centering
\scriptsize
\setlength{\tabcolsep}{2.2pt}
\renewcommand{\arraystretch}{1.08}
\ifdefined\modelhead\else\newcommand{\modelhead}[1]{\rotatebox{60}{\scriptsize\makecell{#1}}}\fi
\caption{Per-model accuracy by reported sub-discipline for all 28 evaluated models, ordered by overall accuracy. Values are percentages; bold indicates the best model for each label. Asterisks on Claude Opus 4.8 (Max) and Claude Opus 5 (Max) denote no-response cases triggered by safety filters on biochemistry questions involving viral biology and pathogen structural characterization; under the strict binary scoring protocol, these instances are counted as incorrect.}\label{tab:accuracy_discipline}
\resizebox{\columnwidth}{!}{%
\begin{tabular}{@{}>{\raggedright\arraybackslash}p{1.55cm}>{\raggedright\arraybackslash}p{4.1cm}cccccccccccccc@{}}
\toprule
\multirow{2}{*}{\makecell[l]{\textbf{Main}\\\textbf{Discipline}}} & \multirow{2}{*}{\textbf{Sub-Discipline}} & \multicolumn{14}{c}{\textbf{Model Accuracy (\%)}} \\
\cmidrule(l){3-16}
 & & \modelhead{gemini-4-argon} & \modelhead{gpt-6-astra-max} & \modelhead{gpt-6.1-sol-max} & \modelhead{GPT-5.6-Sol (Max)} & \modelhead{Gemini 3.1 Pro} & \modelhead{Claude Opus 5\\(Max)*} & \modelhead{kimi-k3.1-pro} & \modelhead{Qwen3.8-Max} & \modelhead{GPT-5.5 (xhigh)} & \modelhead{Kimi K3} & \modelhead{Gemini 3.5 Flash} & \modelhead{grok-4.7} & \modelhead{qwen3.8-flash} & \modelhead{Claude Opus 4.8\\(Max)*} \\
\midrule
\textbf{Chemistry} & Analytical Chemistry & \textbf{56.8} & \textbf{56.8} & 56.6 & 50.5 & 48.2 & 49.5 & 50.2 & 42.3 & 42.5 & 39.5 & 42.5 & 40.9 & 36.6 & 40.9 \\
 & Physical Chemistry & \textbf{51.0} & 44.2 & 45.1 & 41.8 & 46.0 & 42.1 & 37.7 & 31.8 & 35.3 & 32.6 & 39.2 & 32.9 & 32.0 & 31.2 \\
 & Organic Chemistry & \textbf{61.7} & \textbf{61.7} & 60.4 & 53.0 & 50.9 & 53.9 & 51.7 & 46.1 & 46.1 & 40.0 & 47.4 & 43.5 & 42.6 & 49.6 \\
 & Polymer Chemistry and Physics & \textbf{49.4} & 41.4 & 42.0 & 37.9 & 42.5 & 40.8 & 40.2 & 32.2 & 35.1 & 37.4 & 35.1 & 38.5 & 31.0 & 29.9 \\
 & Inorganic Chemistry & 47.3 & 46.4 & \textbf{48.2} & 44.5 & 39.1 & 45.5 & 41.8 & 35.5 & 32.7 & 30.9 & 32.7 & 26.4 & 30.0 & 29.1 \\
\addlinespace
\textbf{Biology} & Biochemistry & \textbf{54.4} & 53.8 & 53.2 & 47.4 & 44.8 & 42.7 & 40.4 & 42.7 & 43.6 & 40.4 & 34.0 & 38.7 & 39.2 & 35.8 \\
 & Molecular Biology & \textbf{57.9} & 55.6 & 54.8 & 48.3 & 44.1 & 39.1 & 44.4 & 47.5 & 43.3 & 42.1 & 32.6 & 35.2 & 39.8 & 34.5 \\
 & Cell Biology & 52.7 & \textbf{55.5} & 54.9 & 53.8 & 43.4 & 39.0 & 41.8 & 48.4 & 46.2 & 44.5 & 34.6 & 37.9 & 42.3 & 36.3 \\
 & Structural Biology & 49.4 & \textbf{54.2} & \textbf{54.2} & 48.8 & 42.2 & 34.3 & 42.8 & 42.2 & 41.0 & 41.0 & 28.3 & 35.5 & 33.1 & 33.1 \\
 & Biophysics & 53.2 & 51.9 & \textbf{53.8} & 47.4 & 44.9 & 38.5 & 40.4 & 36.5 & 38.5 & 39.1 & 37.2 & 35.3 & 36.5 & 30.1 \\
 & Genetics & \textbf{69.4} & \textbf{69.4} & 66.7 & 47.2 & 61.1 & 44.4 & 47.2 & 58.3 & 52.8 & 38.9 & 25.0 & 30.6 & 50.0 & 27.8 \\
 & Immunology & 46.9 & 56.2 & 59.4 & 56.2 & 31.2 & 31.2 & 46.9 & \textbf{62.5} & 43.8 & 43.8 & 34.4 & 40.6 & 40.6 & 34.4 \\
 & Physiology & \textbf{63.2} & 52.6 & 52.6 & 42.1 & 42.1 & 52.6 & 26.3 & 47.4 & 36.8 & 31.6 & 36.8 & 31.6 & 47.4 & 26.3 \\
\addlinespace
\textbf{Materials} & Organic and Polymer Materials & \textbf{46.8} & 40.4 & 41.5 & 37.2 & 39.9 & 34.6 & 34.6 & 25.0 & 30.3 & 29.8 & 32.4 & 33.5 & 22.9 & 27.1 \\
 & Inorganic Non-metallic Materials & \textbf{54.4} & 47.6 & 49.5 & 46.6 & 43.7 & 44.7 & 46.6 & 37.9 & 35.9 & 33.0 & 39.8 & 29.1 & 33.0 & 33.0 \\
 & Composite Materials & \textbf{47.1} & 45.6 & 44.1 & 39.7 & 41.2 & \textbf{47.1} & 39.7 & 29.4 & 32.4 & 35.3 & 35.3 & 30.9 & 30.9 & 32.4 \\
 & Metallic Materials & 55.1 & 55.1 & \textbf{57.1} & \textbf{57.1} & 44.9 & 42.9 & 40.8 & 55.1 & 36.7 & 32.7 & 49.0 & 36.7 & 36.7 & 32.7 \\
\bottomrule
\end{tabular}%
}
\end{table}

%% file: tables/accuracy_by_subdiscipline_model_second.tex
\begin{table}[t]
\centering
\scriptsize
\setlength{\tabcolsep}{2.2pt}
\renewcommand{\arraystretch}{1.08}
\ifdefined\modelhead\else\newcommand{\modelhead}[1]{\rotatebox{60}{\scriptsize\makecell{#1}}}\fi
\caption{Per-model accuracy by reported sub-discipline for all 28 evaluated models, ordered by overall accuracy. Values are percentages; bold indicates the best model for each label. Asterisks on Claude Opus 4.8 (Max) and Claude Opus 5 (Max) denote no-response cases triggered by safety filters on biochemistry questions involving viral biology and pathogen structural characterization; under the strict binary scoring protocol, these instances are counted as incorrect.}\label{tab:accuracy_discipline_second}
\resizebox{\columnwidth}{!}{%
\begin{tabular}{@{}>{\raggedright\arraybackslash}p{1.55cm}>{\raggedright\arraybackslash}p{4.1cm}cccccccccccccc@{}}
\toprule
\multirow{2}{*}{\makecell[l]{\textbf{Main}\\\textbf{Discipline}}} & \multirow{2}{*}{\textbf{Sub-Discipline}} & \multicolumn{14}{c}{\textbf{Model Accuracy (\%)}} \\
\cmidrule(l){3-16}
 & & \modelhead{deepseek-v4.1-flash-1017} & \modelhead{Seed2.1 Pro} & \modelhead{Intern-S2} & \modelhead{Qwen3.7-Plus} & \modelhead{claude-opus-5.5-max} & \modelhead{Intern-S2 Preview-397B} & \modelhead{Seed2.0 Pro} & \modelhead{MiniMax-M3} & \modelhead{Kimi K2.6} & \modelhead{mimo-v2.6-pro} & \modelhead{GLM-5V-Turbo} & \modelhead{Intern-S2 Preview-35B} & \modelhead{MiMo-V2.5} & \modelhead{Intern-S1 Pro} \\
\midrule
\textbf{Chemistry} & Analytical Chemistry & 36.6 & 36.4 & 36.1 & 33.9 & 44.3 & 30.2 & 28.4 & 31.1 & 25.9 & 25.0 & 24.1 & 24.8 & 17.5 & 19.1 \\
 & Physical Chemistry & 28.8 & 25.8 & 31.2 & 27.6 & 37.1 & 28.5 & 21.1 & 27.0 & 24.6 & 25.2 & 19.6 & 20.5 & 13.9 & 16.6 \\
 & Organic Chemistry & 39.6 & 40.9 & 38.7 & 38.7 & 44.8 & 35.2 & 30.9 & 37.0 & 27.0 & 29.6 & 25.7 & 30.0 & 20.4 & 18.7 \\
 & Polymer Chemistry and Physics & 33.3 & 24.1 & 28.7 & 21.8 & 37.4 & 24.7 & 21.3 & 29.9 & 28.2 & 24.7 & 14.9 & 20.7 & 15.5 & 16.7 \\
 & Inorganic Chemistry & 29.1 & 31.8 & 26.4 & 26.4 & 36.4 & 20.0 & 16.4 & 30.0 & 23.6 & 18.2 & 20.9 & 17.3 & 10.0 & 17.3 \\
\addlinespace
\textbf{Biology} & Biochemistry & 34.9 & 33.4 & 32.8 & 33.7 & 12.2 & 29.1 & 32.3 & 23.5 & 23.8 & 25.3 & 22.7 & 18.6 & 16.9 & 14.2 \\
 & Molecular Biology & 36.4 & 36.4 & 30.3 & 35.6 & 10.3 & 28.7 & 33.7 & 21.5 & 24.9 & 23.4 & 23.8 & 18.8 & 19.5 & 12.6 \\
 & Cell Biology & 43.4 & 39.0 & 33.5 & 36.3 & 13.2 & 28.6 & 39.6 & 26.9 & 23.6 & 27.5 & 24.7 & 19.2 & 22.5 & 14.3 \\
 & Structural Biology & 31.9 & 31.3 & 29.5 & 33.1 & 16.3 & 26.5 & 30.1 & 21.1 & 27.1 & 25.3 & 25.3 & 16.9 & 17.5 & 12.7 \\
 & Biophysics & 34.6 & 30.8 & 28.2 & 32.7 & 19.9 & 29.5 & 30.1 & 20.5 & 23.1 & 26.9 & 26.3 & 16.0 & 18.6 & 17.3 \\
 & Genetics & 41.7 & 25.0 & 30.6 & 41.7 & 22.2 & 19.4 & 41.7 & 22.2 & 27.8 & 27.8 & 19.4 & 8.3 & 16.7 & 2.8 \\
 & Immunology & 40.6 & 40.6 & 37.5 & 37.5 & 9.4 & 25.0 & 40.6 & 28.1 & 12.5 & 34.4 & 31.2 & 21.9 & 18.8 & 12.5 \\
 & Physiology & 26.3 & 31.6 & 26.3 & 31.6 & 10.5 & 21.1 & 26.3 & 0.0 & 15.8 & 15.8 & 21.1 & 26.3 & 26.3 & 26.3 \\
\addlinespace
\textbf{Materials} & Organic and Polymer Materials & 28.2 & 22.3 & 25.5 & 17.0 & 37.8 & 20.2 & 14.9 & 25.0 & 26.1 & 22.3 & 15.4 & 18.6 & 16.0 & 13.8 \\
 & Inorganic Non-metallic Materials & 29.1 & 34.0 & 36.9 & 35.0 & 37.9 & 24.3 & 22.3 & 28.2 & 24.3 & 19.4 & 26.2 & 20.4 & 14.6 & 22.3 \\
 & Composite Materials & 25.0 & 23.5 & 36.8 & 22.1 & 42.6 & 32.4 & 25.0 & 27.9 & 25.0 & 19.1 & 14.7 & 23.5 & 13.2 & 14.7 \\
 & Metallic Materials & 30.6 & 36.7 & 30.6 & 36.7 & 51.0 & 34.7 & 30.6 & 34.7 & 26.5 & 22.4 & 26.5 & 32.7 & 12.2 & 14.3 \\
\bottomrule
\end{tabular}%
}
\end{table}

%% file: tables/interdisciplinary_28.tex
\begin{table*}[tp]
\centering\small
\caption{Performance by disciplinary scope across all 28 leaderboard models. Mean accuracy weights models equally. Every question belongs to one main-discipline-scope group and one sub-discipline-count group.}
\label{tab:si-interdisciplinary-28}
\begin{tabular}{lrr}
\toprule
Question group & Questions & Mean accuracy (\%)\\
\midrule
Single main discipline & 602 & 38.6\\
Multiple main disciplines & 514 & 32.0\\
One sub-discipline label & 96 & 39.5\\
Two sub-discipline labels & 416 & 39.3\\
Three sub-discipline labels & 458 & 32.7\\
Four sub-discipline labels & 137 & 30.5\\
Five sub-discipline labels & 9 & 42.1\\
\bottomrule
\end{tabular}
\end{table*}